\documentclass[letterpaper]{article} 
\usepackage{aaai25}  
\usepackage{times}  
\usepackage{helvet}  
\usepackage{courier}  
\usepackage[hyphens]{url}  
\usepackage{graphicx} 
\usepackage{natbib}  
\usepackage{caption} 
\usepackage{algorithm}
\usepackage{algpseudocode}

\usepackage{newfloat}
\usepackage{listings}
\DeclareCaptionStyle{ruled}{labelfont=normalfont,labelsep=colon,strut=off} 
\floatstyle{ruled}
\newfloat{listing}{tb}{lst}{}
\floatname{listing}{Listing}
\usepackage{graphicx}
\usepackage{subcaption}
\usepackage{array}
\usepackage{multirow}
\usepackage{amsthm}
\usepackage{amsmath}
\usepackage{amssymb}
\usepackage{todonotes}
\usepackage{xspace}
\usepackage{marvosym}

\newtheorem{theorem}{Theorem}

\usepackage{threeparttable}
\usepackage{booktabs}

\newcommand{\best}[1]{\textcolor{red}{\textbf{#1}}}
\newcommand{\second}[1]{\textcolor{blue}{\underline{#1}}}
\newcommand{\name}{AMD\xspace}

\usepackage[capitalize]{cleveref}
\crefname{section}{Sec.}{Secs.}
\Crefname{section}{Section}{Sections}
\Crefname{table}{Table}{Tables}
\crefname{table}{Tab.}{Tabs.}

\title{Adaptive Multi-Scale Decomposition Framework for Time Series Forecasting}
\author{
    Yifan Hu\textsuperscript{\rm 1,2,}\equalcontrib, 
    Peiyuan Liu\textsuperscript{\rm 1,}\equalcontrib, 
    Peng Zhu\textsuperscript{\rm 2}, 
    Dawei Cheng\textsuperscript{\rm 2,\thanks{Corresponding author: Dawei Cheng and Tao Dai.}}, 
    Tao Dai\textsuperscript{\rm 3,\footnotemark[2]}
}
\affiliations{
    \textsuperscript{\rm 1}Tsinghua Shenzhen International Graduate School \\
    \textsuperscript{\rm 2}Tongji University \\
    \textsuperscript{\rm 3}Shenzhen University\\

    \text{\{huyf0122, peiyuanliu.edu, daitao.edu\}@gmail.com} \\ \text{\{pengzhu, dcheng\}@tongji.edu.cn}
}

\usepackage{bibentry}

\begin{document}

\maketitle

\begin{abstract}
 Transformer-based and MLP-based methods have emerged as leading approaches in time series forecasting (TSF). 
 However, real-world time series often show different patterns at different scales, and future changes are shaped by the interplay of these overlapping scales, requiring high-capacity models.
 While Transformer-based methods excel in capturing long-range dependencies, they suffer from high computational complexities and tend to overfit. Conversely, MLP-based methods offer computational efficiency and adeptness in modeling temporal dynamics, but they struggle with capturing temporal patterns with complex scales effectively. 
 Based on the observation of multi-scale entanglement effect in time series, we propose a novel MLP-based Adaptive Multi-Scale Decomposition (\name) framework for TSF. Our framework decomposes time series into distinct temporal patterns at multiple scales, leveraging the Multi-Scale Decomposable Mixing (MDM) block to dissect and aggregate these patterns. Complemented by the Dual Dependency Interaction (DDI) block and the Adaptive Multi-predictor Synthesis (AMS) block, our approach effectively models both temporal and channel dependencies and utilizes autocorrelation to refine multi-scale data integration. 
 Comprehensive experiments demonstrate our AMD framework not only overcomes the limitations of existing methods but also consistently achieves state-of-the-art performance across various datasets. 
\end{abstract}

 \begin{links}
     \link{Code}{https://github.com/TROUBADOUR000/AMD}
 \end{links}

%

\section{Introduction}

Time series forecasting (TSF) aims to use historical data to predict future values across various domains, such as finance~\cite{fints, 8425030,lsrigru}, energy~\cite{energy}, traffic management~\cite{traffic}, and weather forecasting~\cite{weather}. Recently, deep learning has made substantial and reliable advancements in TSF, with the most state-of-the-art performances achieved by Transformer-based methods~\cite{patchtst, fedformer, crossformer} and MLP-based methods\cite{dlinear, googlemlpmixer, mtsmixers,duet}.

\begin{figure}
    \centering
    \includegraphics[width=0.465\textwidth]{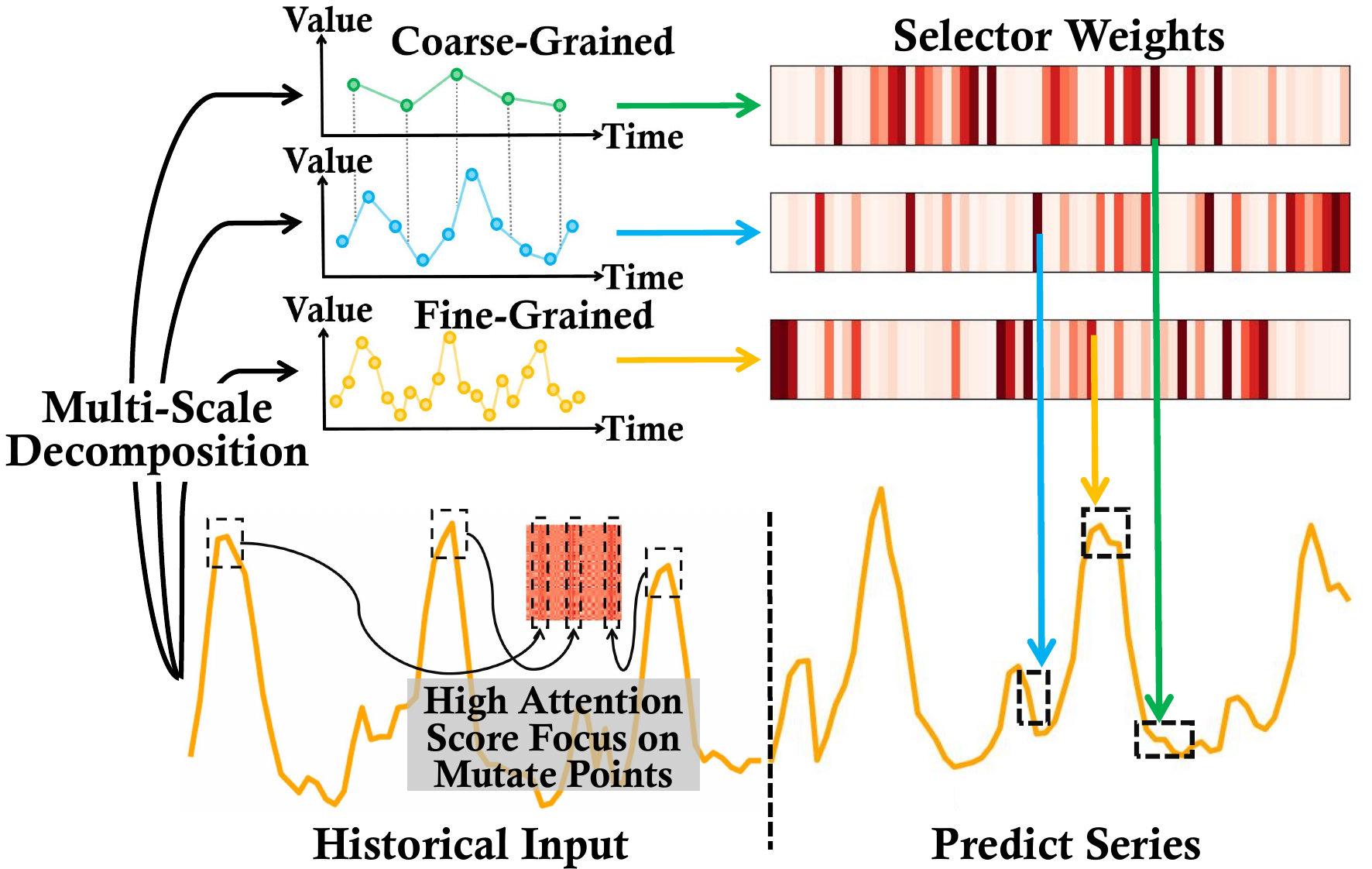} 
    \caption{Illustration of multi-scale temporal patterns in time series and the impact of selector weights. Transformer-based methods often overfit by overemphasizing mutation points, weakening temporal relationships. Efficiently modeling and integrating distinct temporal patterns at various scales is crucial for accurate predictions. 
    }
    \label{fig:intro}
\end{figure}

It is worth noting that time series exhibit distinctly different temporal patterns at various sampling scales~\cite{timemixer}. Moreover, the weight of these time scales in predicting future variations is not uniform, as future variations are jointly determined by the entangle of multiple scales (see \cref{fig:intro}). For example, weather data sampled hourly reflects fine-grained, sudden changes, while monthly sampled data captures coarse-grained climate variations. Similarly, while short-term hourly data might highlight immediate weather shifts, long-term monthly data provides a broader view of climatic trends. 
Therefore, efficiently modeling the multi-scale changes in time series and adeptly integrating information across different scales, which requires high-capacity models, remains a critical challenge.

Although Transformer-based methods excel at modeling long-range dependencies due to the self-attention mechanisms~\cite{vaswani2017attention}, they come with computational complexity that scales quadratically with the length of the sequence. Despite their recognized performance, these methods still face challenges in industrial applications, including low training efficiency and high memory consumption.
Additionally, self-attention can diminish the temporal relationships when extracting semantic correlations between pairs in long sequences~\cite{dlinear}, leading to an overemphasis on mutation points and resulting in overfitting (see \cref{fig:intro}). 
In contrast, MLP-based methods boast significantly lower computational complexity compared to Transformer-based methods. Moreover, MLP-based methods can chronologically model the temporal dynamics in consecutive points, which is crucial for time series analysis~\cite{dlinear, tide}. However, the simplicity of linear mappings in existing MLP-based methods presents an information bottleneck~\cite{kan}, hindering their ability to capture diverse temporal patterns and limiting their predictive accuracy~\cite{linearexperts}.

Motivated by the above observations, we decompose the time series at multiple scales to precisely discern the intertwined temporal patterns within the complex series, rather than merely breaking it down into seasonal and trend components \cite{autoformer, dlinear}. Subsequently, we model the correlations across different scales in both time and channel dimensions. Unlike average aggregation of TimeMixer~\cite{timemixer}, to account for the varying impacts of different temporal patterns on the future, we employ an autocorrelation approach to model their contributions and adaptively integrate these multi-scale temporal patterns based on their respective influences.

Technically, we propose an MLP-based Adaptive Multi-Scale Decomposition (\name) Framework to better disentangle and model the diverse temporal patterns within time series. In concrete, the AMD initiates by employing the Multi-Scale Decomposable Mixing (MDM) block, which first decomposes the original time series into multiple temporal patterns through average downsampling and then aggregates these scales to provide aggregate information in a residual way. Subsequently, the Dual Dependency Interaction (DDI) block simultaneously models both temporal and channel dependencies within the aggregated information. Finally, the Adaptive Multi-predictor Synthesis (AMS) block uses the aggregated information to generate specific weights and then employs these weights to adaptively integrate the multiple temporal patterns produced by the DDI. Through comprehensive experimentation, our AMD consistently achieves state-of-the-art performance in both long-term and short-term forecasting tasks, with superior efficiency 
across multiple datasets. 

Our contributions are summarized as follows: 
\begin{itemize}
\item We decompose time series across multiple scales to precisely identify dominant intertwined temporal patterns within complex sequences and adaptively aggregate predictions of temporal patterns at different scales, addressing their varied impacts on future forecasts. We also demonstrate the feasibility through theoretical analysis. 
\item We propose a simple but effective MLP-based Adaptive Multi-Scale Decomposition (\name) framework that initially decomposes time series into diverse temporal patterns, models both temporal and channel dependencies of these patterns, and finally synthesizes the outputs using a weighted aggregation approach to focus on the changes of dominant temporal patterns, thereby enhancing prediction accuracy across scales. 
\item Comprehensive experiments demonstrate that our AMD consistently delivers state-of-the-art performance in both long-term and short-term forecasting across various datasets, with superior efficiency.
\end{itemize}

\section{Related Works}

\subsection{Time Series Forecasting}
Recently, deep learning methods for TSF have gained prominence 
, such as CNN~\cite{micn, dai2024periodicity}, RNN~\cite{segrnn, WITRANWI}, GNN~\cite{mtgnn, fouriergnn}, Transformer~\cite{autoformer, itransformer} and MLP~\cite{tide, lightts}. 
Transformer-based models, renowned for superior performance in handling long 
sequential data, have gained popularity in TSF. 
PatchTST~\cite{patchtst} divides the time series into patches to enhance locality.
Beyond cross-time dependencies, Crossformer~\cite{crossformer} also mines cross-variable dependencies.
However, Transformer-based models always suffer from efficiency problems due to high computational complexity. In contrast, MLP-based models have a smaller memory footprint.
FITS~\cite{fits} proposes a new linear mapping for the transformation of complex inputs, with only 10k parameters.
However, due to the inherent simplicity and information bottleneck, MLP-based models struggle to effectively capture diverse temporal patterns~\cite{linearexperts}. 
In this work, we decompose time series across multiple scales and use separate linear models for each, effectively addressing such representational limits.


\subsection{Mixture of Experts for TSF}
The concept of Mixture of Experts (MoE) has a long history~\cite{moe,moe1}. Recent advances in large language models (LLM) have reignited interest in MoE~\cite{kim2024scaling, openmoe}. 
However, in the TFS domain, the utilization of MoE remains limited. 
FiLM~\cite{film} assigns input sequences with different time horizons to various experts for forecasting.
FEDformer~\cite{fedformer} extracts a set of data-dependent weights by MoE for combining multiple trend components as the final trend.
In our work, we harness the adaptability of MoE to craft distinct predictors customized for individual temporal patterns, thereby optimizing model performance and enhancing overall model interpretability.

\subsection{Series Decomposition in TSF}
Lately, with high sampling rates leading to high-frequency data (such as daily, hourly, or minutely data), real-world time series data often contains multiple underlying temporal patterns. To competently harness different temporal patterns at various scales, several series decomposition designs are proposed~\cite{pyraformer, nbeats}.
Seasonal-Trend decomposition~\cite{autoformer} uses moving averages to separate seasonal and trend components.
TimesNet~\cite{timesnet} uses Fast Fourier Transform to extract multiple dominant frequencies, while SCINet~\cite{scinet} employs a hierarchical downsampling tree to iteratively decompose multi-scale information.
TimeMixer~\cite{timemixer} downsamples time series into different scales of seasonal and trend sequences.
However, they ignore high-order interactions across different scales.
Refining these designs, we introduces an adaptive multi-scale decomposition method to model the dynamic interactions and accurately discern dominated temporal patterns.


\begin{figure*}
    \centering
    \includegraphics[width=0.932\textwidth]{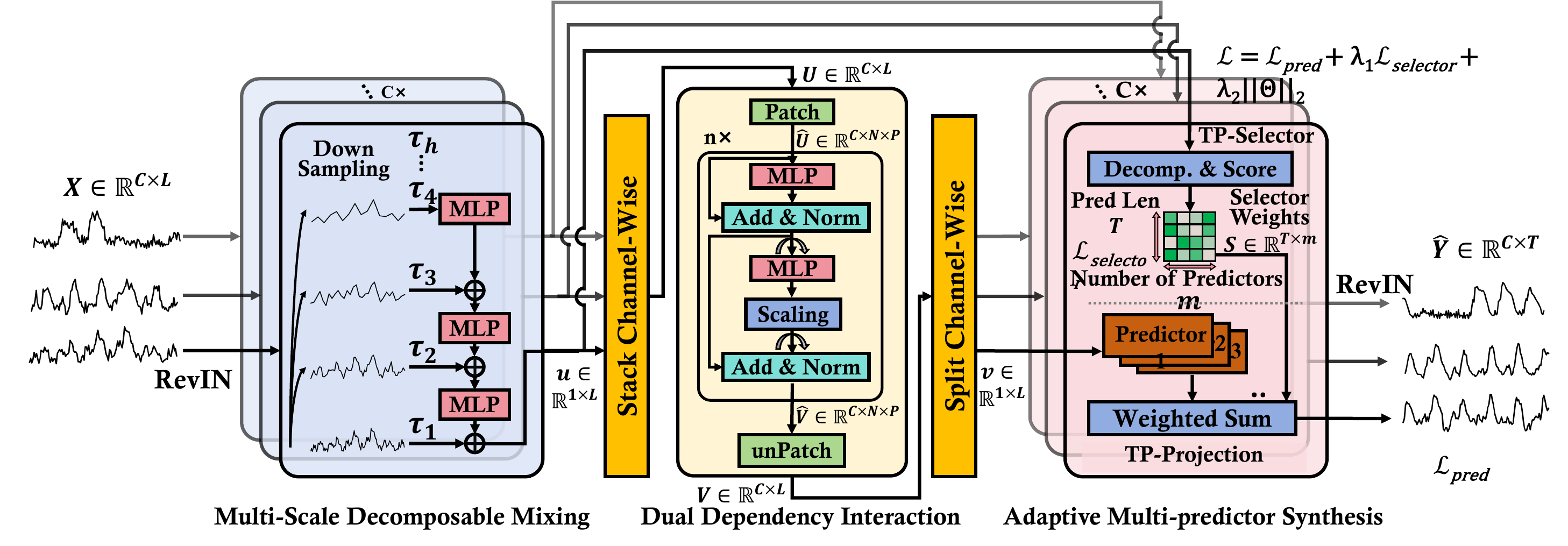} 
    \caption{
    Overall structure of the proposed AMD. MDM decomposes and mixes multi-scale information. DDI models dynamic interactions across different scales. AMS discerns dominated temporal patterns and adaptively makes predictions.
    }
    \label{fig:architecture}
\end{figure*}

\section{Preliminary: Linear Models with Multi-Scale Information for TSF}\label{sec:preliminary}
We consider the following problem: given a collection of time series samples with historical observations $\mathbf{X}\in \mathbb{R}^{C\times L}$, where $C$ denotes the number of variables and $L$ represents the length of the look-back sequence. The objective is to predict $\mathbf{Y}\in \mathbb{R}^{M\times T}$, where $M$ is the number of target variables to be predicted ($M \leq C$) and $T$ is the length of the future time steps to be predicted. A linear model learns parameters $\mathbf{A}\in \mathbb{R}^{L\times T}$ and $\mathbf{b}\in\mathbb{R}^{T}$ to predict the values of next $T$ time steps as:
\begin{equation}
    \mathbf{\hat{Y}} = \mathbf{X}\mathbf{A} \oplus \mathbf{b} \in \mathbb{R}^{C\times L}
\end{equation}
where $\oplus$ means column-wise addition. The corresponding $M$ rows in $\mathbf{\hat{Y}}$ can be used to predict $\mathbf{Y}$.

After that, we introduce the multi-scale information. 
For time series forecasting, the most influential real-world applications typically exhibit either smooth or periodicity. Without these characteristics, predictability tends to be low, rendering predictive models unreliable. 
If the time series only exhibits periodicity, linear models can easily model it~\cite{googletsmixer}. 
We define the original sequence as $f(x)=f_0(x)=[x_1, x_2, ..., x_{L}]$ and assume that $f(x)$ possesses smoothness. After $k$ downsampling operations with a downsampling rate of $d$, we obtain $n$ sequences $f_i(x), \forall i=1,2,...,n$, where $f_i(x)=\frac{1}{d}\sum_{j=xd-d+1}^{xd}f_{i-1}(j)$ and $x=1,2,...,\big[\frac{L}{d^{i}}\big]$. It is noteworthy that $f_i(x) \in \mathbb{R}^{C\times \big[\frac{L}{d^i}\big]}, \forall i=0,1,...,n$. Then, the sequence $f_i(x), \forall i=0,1,...,n$ is transformed into $g_i(x)$ through linear mapping and residual calculation. Specifically, $g_n(x) = f_n(x)$,then through top-down recursion for $i=n-1,..., 0$, the operation 
\begin{equation}
    g_i(x) = f_i(x) + g_{i + 1}(x) W_i
\end{equation}
is performed recursively, where $W_i\in\mathbb{R}^{\big[\frac{L}{d^{i+1}}\big]\times \big[\frac{L}{d^i}\big]}$. In this case, we derive the \textbf{Theorem 1} 

\begin{theorem}
Let multi-scale mixing representation $g(x)$, where $g(x)\in\mathbb{R}^{1\times L}$ (for simplicity, we consider univariate sequences) and the original sequence $f(x)$ is Lipschitz smooth with constant $\mathcal{K}$ ($i.e.\ \ \left| \frac{f(a)-f(b)}{a-b} \right|\leq\mathcal{K}$), then there exists a linear model such that $\left|y_t-\hat{y_t}\right|$ is bounded, $\forall t=1,...,T$.
\end{theorem}

This derivation demonstrates that linear models are well-suited to utilize multi-scale information. For nonperiodic patterns provided they exhibit smoothness, which is often observed in practice, the error remains bounded.

\section{Method}\label{sec:method}

As mentioned above, the key challenge for accurate forecasting is to deal with multi-scale temporal patterns that have an entangled influence on the future evolution of variables. 
In this paper, we propose AMD to enhance forecasting accuracy by capturing multi-scale entanglement through multi-scale decomposition and an autocorrelation approach (see \cref{fig:architecture}).
Specifically, AMD mainly consists of three components: Multi-Scale Decomposable Mixing (MDM) Block, Dual Dependency Interaction (DDI) Block, and Adaptive Multi-predictor Synthesis (AMS) Block. The details of each essential module are explained in the following subsections.

\subsection{Multi-Scale Decomposable Mixing Block}
Time series exhibit both coarse-grained temporal patterns and fine-grained patterns.
Together, these complementary scales of information provide a comprehensive view of the time series. 
Therefore, we first decompose the time series into individual temporal patterns, and then mix them to enhance the time series data for a more nuanced analysis and interpretation.

Specifically, the raw input information $\mathbf{X}$ already contains fine-grained details, while coarse-grained information is extracted through average pooling. 
First-level temporal pattern $\boldsymbol{\tau}_1$ is the input of one channel $\mathbf{x}$. Next, distinct coarse-grained temporal patterns $\boldsymbol{\tau}_i\in\mathbb{R}^{1\times \lfloor \frac{L}{d^{i-1}} \rfloor} (\forall i \in{2,...,h})$ are extracted by applying average pooling over the previous layer of temporal patterns, where $h$ denotes the number of downsampling operations and $d$ denotes the downsampling rate. The decomposition of the $i^{th}$ layer of temporal patterns can be represented as:
\begin{equation}
    \boldsymbol{\tau}_{i}=\text{AvgPooling}(\boldsymbol{\tau}_{i-1})
\end{equation}
Then, distinct temporal patterns are mixed from the coarse-grained $\boldsymbol{\tau}_h$ to the fine-grained $\boldsymbol{\tau}_1$ through a feedforward residual network, while $\xi_i$ represents the mixed data and $\xi_h=\boldsymbol{\tau}_h$.
The mixing of the $i^{th}$ layer of temporal patterns can be represented by the following formula:
\begin{equation}
    \boldsymbol{\xi}_{i}=\boldsymbol{\tau}_{i}+\text{MLP}(\boldsymbol{\xi}_{i+1})
\end{equation}
Finally, after completing the mixing of temporal patterns across $h$ scales, we obtain mixed-scale information $\boldsymbol{\xi}_1$, with the output of one channel being $\mathbf{u}=\boldsymbol{\xi}_{1}\in\mathbb{R}^{1\times L}$.

\subsection{Dual Dependency Interaction Block}
We observed potential dynamic interactions across different scales in various time series. 
Intuitively, in a stock price series, daily fluctuations are influenced by monthly economic trends, which in turn are shaped by annual market cycles. 
As a result, changes at each scale propagate through the system, with the effects at one scale spilling over and altering the dynamics at other scales.
However, ScaleFormer~\cite{scaleformer} and TimeMixer~\cite{timemixer} ignore these potential high-order interactions among different scales.

To model the interactions across different scales, including both temporal and channel dependencies, we propose the DDI block with time-mixing and channel-mixing.
DDI first stacks the aggregated information $\mathbf{u}$ from various channels of the MDM into the matrix $\mathbf{U} \in \mathbb{R}^{C \times L}$ and then performs patch operations to transform $\mathbf{U}$ into $\hat{\mathbf{U}} \in \mathbb{R}^{C \times N \times P}$. 
For each patch, $\hat{\mathbf{V}}_{t}^{t+P}$ represents the embedding output of the residual network, and $\hat{\mathbf{U}}_{t}^{t+P}$ represents a patch of aggregated information from MDM. 
We adopt an MLP shared by time steps to aggregate time-mixing information through the time dimension for each channel to obtain temporal dependencies $\mathbf{Z}_{t}^{t+P}$. 
Next, we perform the transpose operation to fuse channel-mixing information along the time domain through another MLP shared by channels. Finally, we perform the unpatch operation and split the output information into individual channels to obtain $\mathbf{v}\in\mathbb{R}^{1\times L}$. 
The residual operation ensures that the model retains its ability to capture temporal dependencies while effectively exploiting cross-channel dependencies.
The interaction of the patch $\hat{\mathbf{U}}_{t}^{t+P}$ can be represented by the following formula:
\begin{equation}
    \mathbf{Z}_{t}^{t+P} = \hat{\mathbf{U}}_{t}^{t+P} + \text{MLP}(\hat{\mathbf{V}}_{t-P}^{t})
\end{equation}
\begin{equation}
    \hat{\mathbf{V}}_{t}^{t+P} = \mathbf{Z}_{t}^{t+P} + \beta\cdot \text{MLP}((\mathbf{Z}_{t}^{t+P})^T)^T
\end{equation}
where $A^T$ is the transpose of matrix A. 

In DDI, dual dependencies are captured under the mixed-scale information $\mathbf{U}$. Temporal dependencies model the interactions across different periods, while cross-channel dependencies model the relationships between different variables.
However, through experiments, we find that cross-channel dependencies are not always effective, especially when the target time series is not correlated with other covariates; instead, they often introduce unwanted interference. Therefore, we introduce the scaling rate $\beta$ to suppress the noise and balance the emphasis on temporal dependencies and cross-channel dependencies.

\subsection{Adaptive Multi-predictor Synthesis Block}
It is noted that the dominant temporal patterns, which significantly influence future variables, dynamically change over different periods and have stronger predictive power. In contrast to the average aggregation in TimeMixer~\cite{timemixer}, which overlooks these dominant changes, we exploit the adaptive properties of MoE to design specific predictors for each temporal pattern. By dynamically assigning more attention to the dominant scales, we improve both accuracy and generalisability.

The AMS is partitioned into two components: the temporal pattern selector (TP-Selector) and the temporal pattern projection (TP-Projection). The TP-Selector decomposes different temporal patterns and generates the selector weights $\mathbf{S}$. Unlike the downsampling in MDM, which decomposes individual scales to enhance temporal information, the TP-Selector adaptively untangles highly correlated, intertwined mixed scales through feedforward processing. Meanwhile, the TP-Projection synthesizes the multi-predictions and adaptively aggregates the outputs based on the specific weights.

\textbf{TP-Selector} takes a single channel input $\mathbf{u}$ from MDM, decomposes it through feedforward layers, and then applies a noisy gating design~\cite{llm}:
\begin{equation}
    \mathbf{S}=\text{Softmax}(\text{TopK}(\text{Softmax}(Q(\mathbf{u})), k))
    \label{S_eq}
\end{equation}
\begin{equation}
    Q(\mathbf{u})=\text{Decomp.}(\mathbf{u})+\psi\cdot \text{Softplus}(\text{Decomp.}(\mathbf{u})\cdot \mathbf{W}_\text{noise})
\end{equation}
where $k$ is the number of dominant temporal patterns, $\psi \in \mathbb{N}(0,1)$ is standard Gaussian noise, and $\mathbf{W}_\text{noise} \in \mathbb{R}^{m \times m}$ is a learnable weight controlling the noisy values. 

\renewcommand{\arraystretch}{0.61}
\begin{table*}[htb]
\setlength{\tabcolsep}{3.7pt}
\scriptsize
\centering
\begin{threeparttable}
\begin{tabular}{c|c|cc|cc|cc|cc|cc|cc|cc|cc|cc|cc}
\toprule

\multicolumn{2}{c}{\scalebox{1.1}{Models}} & \multicolumn{2}{c}{AMD (Ours)} & \multicolumn{2}{c}{TimeMixer} & \multicolumn{2}{c}{PatchTST} & \multicolumn{2}{c}{iTransformer} & \multicolumn{2}{c}{Crossformer} & \multicolumn{2}{c}{FEDformer} & \multicolumn{2}{c}{TimesNet} & \multicolumn{2}{c}{MICN} & \multicolumn{2}{c}{DLinear} & \multicolumn{2}{c}{MTS-Mixers} \\ 



 \cmidrule(lr){3-4} \cmidrule(lr){5-6} \cmidrule(lr){7-8} \cmidrule(lr){9-10} \cmidrule(lr){11-12} \cmidrule(lr){13-14} \cmidrule(lr){15-16} \cmidrule(lr){17-18} \cmidrule(lr){19-20} \cmidrule(lr){21-22} 

\multicolumn{2}{c}{Metric} & MSE & MAE & MSE & MAE & MSE & MAE & MSE & MAE & MSE & MAE & MSE & MAE & MSE & MAE & MSE & MAE & MSE & MAE & MSE & MAE \\
 
\toprule

\multirow{4}{*}{\rotatebox{90}{Weather}}
& 96 & \best{0.145} & \best{0.197} & \second{0.147} & \best{0.197} & 0.149 & \second{0.198} & 0.174 & 0.214 & 0.153 & 0.217 & 0.238 & 0.314 & 0.172 & 0.220 & 0.161 & 0.226 & 0.152 & 0.237 & 0.156 & \second{0.206} \\
& 192 & \best{0.187} & \best{0.238} & \second{0.189} & \second{0.239} & 0.194 & 0.241 & 0.221 & 0.254 & 0.197 & 0.269 & 0.275 & 0.329 & 0.219 & 0.261 & 0.220 & 0.283 & 0.220 & 0.282 & 0.199 & 0.248 \\
& 336 & \best{0.240} & \best{0.280} & \second{0.241} & \best{0.280} & 0.245 & 0.282 & 0.278 & 0.296 & 0.252 & 0.311 & 0.339 & 0.377 & 0.280 & 0.306 & 0.275 & 0.328 & 0.265 & 0.319 & 0.249 & 0.291 \\
& 720 & 0.315 & \best{0.330} & \best{0.310} & \best{0.330} & 0.314 & \second{0.334} & 0.358 & 0.349 & 0.318 & 0.363 & 0.389 & 0.409 & 0.365 & 0.359 & \second{0.311} & 0.356 & 0.323 & 0.362 & 0.336 & 0.343 \\

\midrule

\multirow{4}{*}{\rotatebox{90}{ETTh1}}
& 96 & \second{0.369} & \second{0.397} & \best{0.361} & \best{0.390} & 0.370 & 0.399 & 0.386 & 0.405 & 0.386 & 0.429 & 0.376 & 0.415 & 0.384 & 0.402 & 0.396 & 0.427 & 0.375 & \second{0.399} & 0.372 & 0.395 \\
& 192 & \best{0.401} & \second{0.416} & 0.409 & \best{0.414} & 0.413 & 0.421 & 0.441 & 0.436 & 0.416 & 0.444 & 0.423 & 0.446 & 0.457 & 0.436 & 0.430 & 0.453 & \second{0.405} & \second{0.416} & 0.416 & 0.426 \\
& 336 & \best{0.418} & \best{0.427} & 0.430 & \second{0.429} & \second{0.422} & 0.436 & 0.487 & 0.458 & 0.440 & 0.461 & 0.444 & 0.462 & 0.491 & 0.469 & 0.433 & 0.458 & 0.439 & 0.443 & 0.455 & 0.449 \\
& 720 & \best{0.439} & \best{0.454} & \second{0.445} & \second{0.460} & 0.447 & 0.466 & 0.503 & 0.491 & 0.519 & 0.524 & 0.469 & 0.492 & 0.521 & 0.500 & 0.474 & 0.508 & 0.472 & 0.490 & 0.475 & 0.472 \\

\midrule

\multirow{4}{*}{\rotatebox{90}{ETTh2}}
& 96 & \second{0.274} & \second{0.337} & \best{0.271} & \best{0.330} & \second{0.274} & \second{0.337} & 0.297 & 0.349 & 0.628 & 0.563 & 0.332 & 0.374 & 0.340 & 0.374 & \second{0.289} & 0.357 & 0.289 & 0.353 & 0.307 & 0.354 \\
& 192 & 0.351 & 0.383 & \best{0.317} & \best{0.402} & \second{0.339} & \second{0.379} & 0.380 & 0.400 & 0.703 & 0.624 & 0.407 & 0.446 & 0.402 & 0.414 & 0.409 & 0.438 & 0.383 & 0.418 & 0.374 & 0.399 \\
& 336 & 0.375 & 0.411 & \second{0.332} & \second{0.396} & \best{0.329} & \best{0.380} & 0.428 & 0.432 & 0.827 & 0.675 & 0.400 & 0.447 & 0.452 & 0.452 & 0.417 & 0.452 & 0.448 & 0.465 & 0.398 & 0.432 \\
& 720 & 0.402 & 0.438 & \best{0.342} & \best{0.408} & \second{0.379} & \second{0.422} & 0.427 & 0.445 & 1.181 & 0.840 & 0.412 & 0.469 & 0.462 & 0.468 & 0.426 & 0.473 & 0.605 & 0.551 & 0.463 & 0.465 \\

\midrule

\multirow{4}{*}{\rotatebox{90}{ETTm1}}
& 96 & \best{0.284} & \best{0.339} & 0.291 & \second{0.340} & \second{0.290} & 0.342 & 0.334 & 0.368 & 0.316 & 0.373 & 0.326 & 0.390 & 0.338 & 0.375 & 0.314 & 0.360 & 0.299 & 0.343 & 0.314 & 0.358 \\
& 192 & \best{0.322} & \best{0.362} & \second{0.327} & \second{0.365} & 0.332 & 0.369 & 0.377 & 0.391 & 0.377 & 0.411 & 0.365 & 0.415 & 0.371 & 0.387 & 0.359 & 0.387 & 0.335 & \second{0.365} & 0.354 & 0.386 \\
& 336 & \best{0.360} & \best{0.380} & \best{0.360} & \second{0.381} & \second{0.366} & 0.392 & 0.426 & 0.420 & 0.431 & 0.442 & 0.391 & 0.425 & 0.410 & 0.411 & 0.398 & 0.413 & 0.369 & 0.386 & 0.384 & 0.405 \\
& 720 & 0.421 & \best{0.416} & \best{0.415} & \second{0.417} & \second{0.416} & 0.420 & 0.491 & 0.459 & 0.600 & 0.547 & 0.446 & 0.458 & 0.478 & 0.450 & 0.459 & 0.464 & 0.425 & 0.421 & 0.427 & 0.432 \\

\midrule

\multirow{4}{*}{\rotatebox{90}{ETTm2}}
& 96 & 0.167 & 0.258 & \best{0.164} & \best{0.254} & \second{0.166} & \second{0.256} & 0.180 & 0.264 & 0.421 & 0.461 & 0.180 & 0.271 & 0.187 & 0.267 & 0.178 & 0.273 & \second{0.167} & 0.260 & 0.177 & 0.259 \\
& 192 & \best{0.221} & \best{0.294} & \second{0.223} & \second{0.295} & \second{0.223} & 0.296 & 0.250 & 0.309 & 0.503 & 0.519 & 0.252 & 0.318 & 0.249 & 0.309 & 0.245 & 0.316 & 0.224 & 0.303 & 0.241 & 0.303 \\
& 336 & \best{0.270} & \best{0.327} & 0.279 & 0.330 & \second{0.274} & \second{0.329} & 0.311 & 0.348 & 0.611 & 0.580 & 0.324 & 0.364 & 0.321 & 0.351 & 0.295 & 0.350 & 0.281 & 0.342 & 0.297 & 0.338 \\
& 720 & \best{0.356} & \best{0.382} & \second{0.359} & \second{0.383} & 0.362 & 0.385 & 0.412 & 0.407 & 0.996 & 0.750 & 0.410 & 0.420 & 0.497 & 0.403 & 0.389 & 0.409 & 0.397 & 0.421 & 0.396 & 0.398 \\

\midrule

\multirow{4}{*}{\rotatebox{90}{ECL}}
& 96 & \best{0.129} & \second{0.224} & \best{0.129} & \second{0.224}  & \best{0.129} & \best{0.222} & 0.148 & 0.240 & 0.187 & 0.283 & 0.186 & 0.302 & 0.168 & 0.272 & 0.159 & 0.267 & 0.153 & 0.237 & \second{0.141} & 0.243 \\
& 192 & \second{0.147} & \second{0.238} & \best{0.140} & \best{0.220} & \second{0.147} & 0.240 & 0.162 & 0.253 & 0.258 & 0.330 & 0.197 & 0.311 & 0.184 & 0.289 & 0.168 & 0.279 & 0.152 & 0.249 & 0.163 & 0.261 \\
& 336 & \best{0.160} & \best{0.253} & \second{0.161} & \second{0.255} & 0.163 & 0.259 & 0.178 & 0.269 & 0.323 & 0.369 & 0.213 & 0.328 & 0.198 & 0.300 & 0.196 & 0.308 & 0.169 & 0.267 & 0.176 & 0.277 \\
& 720 & \best{0.193} & \best{0.286} & \second{0.194} & \second{0.287} & 0.197 & 0.290 & 0.225 & 0.317 & 0.404 & 0.423 & 0.233 & 0.344 & 0.220 & 0.320 & 0.203 & 0.312 & 0.233 & 0.344 & 0.212 & 0.308 \\

\midrule

\multirow{4}{*}{\rotatebox{90}{Exchange}}
& 96 & \second{0.083} & \best{0.201} & 0.086 & 0.204 & 0.093 & 0.214 & 0.086 & 0.206 & 0.186 & 0.346 & 0.136 & 0.276 & 0.107 & 0.234 & 0.102 & 0.235 & \best{0.081} & \second{0.203} & 0.083 & 0.201 \\
& 192 & \second{0.171} & \best{0.293} & 0.176 & 0.298 & 0.192 & 0.312 & 0.177 & 0.299 & 0.467 & 0.522 & 0.256 & 0.369 & 0.226 & 0.344 & 0.172 & 0.316 & \best{0.157} & \best{0.293} & 0.174 & \second{0.296} \\
& 336 & 0.309 & \best{0.402} & 0.345 & 0.426 & 0.350 & 0.432 & 0.331 & 0.417 & 0.783 & 0.721 & 0.426 & 0.464 & 0.367 & 0.448 & \best{0.272} & \second{0.407} & \second{0.305} & 0.414 & 0.336 & 0.417 \\
& 720 & 0.750 & \second{0.652} & 0.848 & 0.692 & 0.911 & 0.716 & 0.847 & 0.691 & 1.367 & 0.943 & 1.090 & 0.800 & 0.964 & 0.746 & \second{0.714} & 0.658 & \best{0.643} & \best{0.601} & 0.900 & 0.715 \\

\midrule

\multirow{4}{*}{\rotatebox{90}{Traffic}}
& 96 & \second{0.366} & \second{0.259} & \best{0.360} & \best{0.249} & \best{0.360} & \best{0.249} & 0.395 & 0.268 & 0.512 & 0.290 & 0.576 & 0.359 & 0.593 & 0.321 & 0.508 & 0.301 & 0.410 & 0.282 & 0.462 & 0.332 \\
& 192 & 0.381 & 0.265 & \second{0.379} & \second{0.256} & \best{0.375} & \best{0.350} & 0.417 & 0.276 & 0.523 & 0.297 & 0.610 & 0.380 & 0.617 & 0.336 & 0.536 & 0.315 & 0.423 & 0.287 & 0.488 & 0.354 \\
& 336 & 0.397 & \second{0.269} & \best{0.385} & 0.270 & \second{0.392} & \best{0.264} & 0.433 & 0.283 & 0.530 & 0.300 & 0.608 & 0.375 & 0.629 & 0.336 & 0.525 & 0.310 & 0.436 & 0.296 & 0.498 & 0.360 \\
& 720 & \best{0.429} & 0.292 & \second{0.430} & \best{0.281} & 0.432 & \second{0.286} & 0.467 & 0.302 & 0.573 & 0.313 & 0.621 & 0.375 & 0.640 & 0.350 & 0.571 & 0.323 & 0.466 & 0.315 & 0.529 & 0.370 \\

\midrule

\multirow{4}{*}{\rotatebox{90}{Solar}}
& 96 & \second{0.175} & \second{0.228} & \best{0.167} & \best{0.220} & 0.224 & 0.278 & 0.203 & 0.237 & 0.181 & 0.240 & 0.201 & 0.304 & 0.219 & 0.314 & 0.188 & 0.252 & 0.289 & 0.337 & 0.284 & 0.325 \\
& 192 & \best{0.186} & \best{0.235} & \second{0.187} & \second{0.249} & 0.253 & 0.298 & 0.233 & 0.261 & 0.196 & 0.252 & 0.237 & 0.337 & 0.231 & 0.322 & 0.215 & 0.280 & 0.319 & 0.397 & 0.307 & 0.362 \\
& 336 & \best{0.200} & \second{0.246} & \best{0.200} & 0.258 & 0.273 & 0.306 & 0.248 & 0.273 & \second{0.216} & \best{0.243} & 0.254 & 0.362 & 0.246 & 0.337 & 0.222 & 0.267 & 0.352 & 0.415 & 0.333 & 0.384 \\
& 720 & \best{0.203} & \best{0.249} & \second{0.215} & \second{0.250} & 0.272 & 0.308 & 0.249 & 0.275 & 0.220 & 0.256 & 0.280 & 0.397 & 0.280 & 0.363 & 0.226 & 0.264 & 0.356 & 0.412 & 0.335 & 0.383 \\



\bottomrule

\end{tabular}
\caption{Long-term forecasting task. All the results are averaged from 4 different prediction lengths $T\in\{96, 192, 336, 720\}$. To the best for a fairer comparison for all baselines, the input sequence length $L$ is searched among $\{96, 192, 336, 512, 672, 720\}$.}
\label{tab:long}
\end{threeparttable}
\end{table*}
\renewcommand{\arraystretch}{0.61}

\textbf{TP-Projection} takes the embedding $\mathbf{v}$ from the output $\mathbf{V}$ of DDI as input. Each predictor consists of multiple feedforward layers. These outputs are then multiplied by the $\mathbf{S}$ and summed to yield the prediction result $\mathbf{\hat{y}}$ for one channel. The final results $\mathbf{\hat{Y}}$ are composed of the outputs $\mathbf{\hat{y}}$ from each channel:

\begin{equation}
    \mathbf{\hat{y}}=\sum_{j=0}^{m} \mathbf{S}_j\cdot \text{Predictor}_j(\mathbf{v})
\end{equation}

Compared to sparse MoE~\cite{sparse}, we adopt dense MoE in TP-Projection with two considerations. 
Firstly, each temporal pattern contributes to the prediction result. 
Secondly, this approach can help mitigate issues such as load unbalancing and embedding omissions that are prevalent in sparse MoE architectures. 
Our $\text{TopK}$ method can be formalized with:
\begin{equation}
    \text{TopK}(\mathbf{u}, k)=
    \begin{cases}
        \alpha\cdot \mathrm{log}(\mathbf{u}+1), \ \ \text{if}\ \mathbf{u} < v_k\\
        \alpha\cdot \mathrm{exp}(\mathbf{u})-1, \ \ \text{if}\ \mathbf{u} \geq v_k
    \end{cases}
\end{equation}
where $v_k$ is the k-th largest value among $\mathbf{u}$ and $\alpha$ is a constant used to adjust the selector weights. The scaling operation within our TopK requires the input values to be restricted to the interval [0,1]. Consequently, we perform an additional $\text{Softmax}$ operation according to \cref{S_eq}.

\subsection{Loss Function}

The loss function of AMS consists of two components. (1) For the predictors, the Mean Squared Error (MSE) loss ($\boldsymbol{\mathcal{L}}_\text{pred}=\sum_{i=0}^{T}\|\mathbf{y_i-\hat{y_i}}\|_{2}^{2}$) is used to measure the variance between predicted values and ground truth. (2) For the gating network loss, we apply the coefficient of variation loss function ($\boldsymbol{\mathcal{L}}_\text{selector}=\frac{Var(S)}{Mean(S)^2+\epsilon}$, where $\epsilon$ is a small positive constant to prevent numerical instability), which optimizes the gating mechanism by promoting a balanced assignment of experts to inputs, thereby enhancing the overall performance of the MoE. The total loss function is defined as:
\begin{equation}
\boldsymbol{\mathcal{L}}=\boldsymbol{\mathcal{L}}_\text{pred}+\lambda_{1}\boldsymbol{\mathcal{L}}_\text{selector}+\lambda_{2}\|\mathbf{\Theta}\|_{2}
\label{eq:loss}
\end{equation}
where $\|\mathbf{\Theta}\|_{2}$ is the L2-norm, $\lambda_{1,2}$ are hyper-parameters.

\section{Experiments}\label{sec:experiment}

\subsection{Main Results}

\paragraph{Datasets.}
We conduct experiments on seven real-world datasets, including Weather, ETT (ETTh1, ETTh2, ETTm1, ETTm2), ECL, Exchange, Traffic and Solar Energy for long-term forecasting and PEMS (PEMS03, PEMS04, PEMS07, PEMS08) for short-term forecasting. 

\paragraph{Baselines.}
We carefully select some representative models to serve as baselines in the field of time series forecasting, including (1) MLP-based models: TimeMixer~\cite{timemixer}, TiDE~\cite{tide}, MTS-Mixers~\cite{mtsmixers}, RLinear~\cite{rlinear}, and DLinear~\cite{dlinear}; (2) Transformer-based models: PatchTST~\cite{patchtst}, iTransformer~\cite{itransformer}, Crossformer~\cite{crossformer}, and FEDformer~\cite{fedformer}; (3) CNN-based models: TimesNet~\cite{timesnet}, and MICN~\cite{micn}. 

\paragraph{Experimental Settings.}
To ensure fair comparisons, for long-term forecasting, we rerun all baselines with different input lengths $L$ and choose the best results to avoid underestimating the baselines. For short-term forecasting, the input length is 96.
We select two common metrics in time series forecasting: Mean Absolute Error (MAE) and Mean Squared Error (MSE). 
All experiments are conducted using PyTorch on an NVIDIA V100 32GB GPU and are repeated five times for consistency.

\renewcommand{\arraystretch}{0.3}
\begin{table*}[htb]
\setlength{\tabcolsep}{4pt}
\scriptsize
\centering
\begin{threeparttable}
\begin{tabular}{c|c|cc|cc|cc|cc|cc|cc|cc|cc|cc}
\toprule

\multicolumn{2}{c}{\scalebox{1.1}{Models}} & \multicolumn{2}{c}{AMD} & \multicolumn{2}{c}{PatchTST} & \multicolumn{2}{c}{Crossformer} & \multicolumn{2}{c}{FEDformer} & \multicolumn{2}{c}{TimesNet} & \multicolumn{2}{c}{TiDE} & \multicolumn{2}{c}{DLinear} & \multicolumn{2}{c}{MTS-Mixers} & \multicolumn{2}{c}{RLinear} \\ 


 \cmidrule(lr){3-4} \cmidrule(lr){5-6} \cmidrule(lr){7-8} \cmidrule(lr){9-10} \cmidrule(lr){11-12} \cmidrule(lr){13-14} \cmidrule(lr){15-16} \cmidrule(lr){17-18} \cmidrule(lr){19-20} 

\multicolumn{2}{c}{Metric} & MSE & MAE & MSE & MAE & MSE & MAE & MSE & MAE & MSE & MAE & MSE & MAE & MSE & MAE & MSE & MAE & MSE & MAE \\
 
\toprule

\multicolumn{2}{c|}{\scalebox{0.9}{PEMS03}} & \best{0.084} & \second{0.198} & 0.099 & 0.216 & 0.090 & 0.203 & 0.126 & 0.251 & \second{0.085} & \best{0.192} & 0.178 & 0.305 & 0.122 & 0.243 & 0.117 & 0.232 & 0.126 & 0.236 \\

\midrule

\multicolumn{2}{c|}{\scalebox{0.9}{PEMS04}} & \best{0.083} & \second{0.198} & 0.105 & 0.224 & 0.098 & 0.218 & 0.138 & 0.262 & \second{0.087} & \best{0.195} & 0.219 & 0.340 & 0.148 & 0.272 & 0.129 & 0.267 & 0.138 & 0.252 \\

\midrule

\multicolumn{2}{c|}{\scalebox{0.9}{PEMS07}} & \best{0.074} & \best{0.180} & 0.095 & 0.150 & 0.094 & 0.200 & 0.109 & 0.225 & \second{0.082} & \second{0.181} & 0.173 & 0.304 & 0.115 & 0.242 & 0.134 & 0.278 & 0.118 & 0.235 \\

\midrule

\multicolumn{2}{c|}{\scalebox{0.9}{PEMS08}} & \best{0.093} & \best{0.206} & 0.168 & 0.232 & 0.165 & 0.214 & 0.173 & 0.273 & \second{0.112} & \second{0.212} & 0.227 & 0.343 & 0.154 & 0.276 & 0.186 & 0.286 & 0.133 & 0.247 \\

\bottomrule

\end{tabular}
\caption{Short-term forecasting task. The input sequence length $L$ is 96 and the prediction length $T$ is 12.}
\label{tab:short}
\end{threeparttable}
\end{table*}
\renewcommand{\arraystretch}{0.3}

\renewcommand{\arraystretch}{0.5}
\begin{table*}[htb]
\setlength{\tabcolsep}{3pt}
\scriptsize
\centering
\begin{threeparttable}
\begin{tabular}{c|c|cc|cc|cc|cc|cc|cc|cc|cc|cc|cc}
\toprule

\multicolumn{2}{c}{\scalebox{1.1}{Models}} & \multicolumn{2}{c}{AMD (Ours)} & \multicolumn{2}{c}{AMD (Sparse)} & \multicolumn{2}{c}{\textbf{RandomOrder}} & \multicolumn{2}{c}{\textbf{AverageWeight}} & \multicolumn{2}{c}{\textbf{w/o DDI}} & \multicolumn{2}{c}{$\beta=0.5$} & \multicolumn{2}{c}{$\beta=1.0$} & \multicolumn{2}{c}{\textbf{w/o} $\boldsymbol{\mathcal{L}}_\text{selector}$} & \multicolumn{2}{c}{\textbf{w/o MDM}} & \multicolumn{2}{c}{\textbf{w/o AMS}} \\

 \cmidrule(lr){3-4} \cmidrule(lr){5-6} \cmidrule(lr){7-8} \cmidrule(lr){9-10} 
 \cmidrule(lr){11-12} \cmidrule(lr){13-14} \cmidrule(lr){15-16} \cmidrule(lr){17-18} \cmidrule(lr){19-20} \cmidrule(lr){21-22} 

\multicolumn{2}{c}{Metric} & MSE & MAE & MSE & MAE & MSE & MAE & MSE & MAE & MSE & MAE & MSE & MAE & MSE & MAE & MSE & MAE & MSE & MAE & MSE & MAE \\
 
\toprule

\multirow{4}{*}{\rotatebox{90}{Weather}}
& 96  & \best{0.145} & \best{0.198} & 0.150 & 0.205 & 0.152 & 0.209 & 0.149 & 0.205 & 0.154 & 0.208 & 0.146 & \best{0.198} & 0.147 & 0.200 & 0.160 & 0.211 & 0.149 & 0.204 & 0.150 & 0.206 \\
& 192 & \best{0.187} & \best{0.238} & 0.193 & 0.245 & 0.194 & 0.245 & 0.194 & 0.246 & 0.197 & 0.248 & 0.194 & 0.245 & 0.195 & 0.246 & 0.209 & 0.258 & 0.192 & 0.243 & 0.199 & 0.248 \\
& 336 & \best{0.240} & \best{0.280} & 0.244 & 0.284 & 0.245 & 0.286 & 0.249 & 0.288 & 0.249 & 0.289 & 0.249 & 0.287 & 0.248 & 0.286 & 0.278 & 0.309 & 0.246 & 0.285 & 0.248 & 0.287 \\
& 720 & \best{0.315} & \best{0.330} & \best{0.315} & 0.334 & 0.321 & 0.338 & 0.325 & 0.340 & 0.323 & 0.340 & 0.318 & 0.335 & 0.320 & 0.336 & 0.364 & 0.366 & 0.319 & 0.336 & 0.320 & 0.339 \\

\midrule

\multirow{4}{*}{\rotatebox{90}{ECL}}
& 96  & \best{0.129} & \best{0.224} & 0.132 & 0.228 & 0.133 & 0.230 & 0.135 & 0.234 & 0.132 & 0.231 & 0.140 & 0.242 & 0.147 & 0.248 & 0.139 & 0.238 & 0.135 & 0.231 & 0.137 & 0.234 \\
& 192 & \best{0.147} & \best{0.238} & 0.151 & 0.244 & 0.155 & 0.248 & 0.155 & 0.251 & 0.151 & 0.247 & 0.169 & 0.268 & 0.175 & 0.273 & 0.156 & 0.456 & 0.153 & 0.247 & 0.157 & 0.251 \\
& 336 & \best{0.160} & \best{0.253} & 0.165 & 0.259 & 0.171 & 0.264 & 0.169 & 0.263 & 0.167 & 0.266 & 0.174 & 0.274 & 0.183 & 0.285 & 0.177 & 0.271 & 0.169 & 0.265 & 0.171 & 0.267 \\
& 720 & \best{0.193} & \best{0.286} & 0.199 & 0.293 & 0.202 & 0.294 & 0.202 & 0.295 & 0.200 & 0.294 & 0.207 & 0.298 & 0.215 & 0.308 & 0.208 & 0.299 & 0.201 & 0.296 & 0.203 & 0.298 \\

\bottomrule

\end{tabular}
\caption{Component ablation of AMD on Weather and ECL.}
\label{tab:ablation}
\end{threeparttable}
\end{table*}
\renewcommand{\arraystretch}{0.5}

\paragraph{Results.}
Comprehensive forecasting results are shown in \cref{tab:long} and \cref{tab:short}, which present long-term and short-term forecasting results respectively. The best results are highlighted in \textcolor{red}{\textbf{red}} and the second-best are \textcolor{blue}{\underline{underlined}}.
The lower the MSE/MAE, the more accurate the forecast. AMD stands out with the best performance in 50 cases and the second best in 27 cases out of the overall 80 cases. Compared with other baselines, AMD performs well on both high-dimensional and low-dimensional datasets. It is worth noting that PatchTST does not perform well on the PEMS datasets for the patching design leads to the neglect of highly fluctuating temporal patterns. In contrast, AMD leverages information from multi-scale temporal patterns. Furthermore, the performance of Crossformer is unsatisfactory for it introduces unnecessary noise by exploring cross-variable dependencies. AMD skillfully reveals the intricate dependencies existing among time steps across various variables. Additionally, DLinear and MTS-mixers perform poorly on high-dimensional datasets, whereas AMD can handle them. 


\subsection{Model Analysis}


\paragraph{Ablation Study.}
To verify the effectiveness of each component of AMD, we perform detailed ablation
of each possible design on Weather and ECL datasets.
As shown in \cref{tab:ablation}, we have the following observations.

For MDM, we conduct experiments by removing MDM (\textbf{w/o MDM}). The increased error suggests that relying on a single scale is insufficient for accurate prediction.

For DDI, recent studies~\cite{patchtst, crossformer} show that both channel independence and cross-channel dependencies strategies can achieve SOTA accuracy in specific tasks. Typically, datasets with strong cross-channel correlations perform better using cross-channel dependencies strategies. However, for datasets with weak inter-channel correlations, cross-channel dependencies often introduce unwanted noise. 
To prove this, we introduce cross-channel dependencies by adjusting the scaling rate. Specifically, we set the parameter $\beta$ in \cref{S_eq} to 0.5 and 1.0, respectively. In addition, we conduct experiments by removing DDI (\textbf{w/o DDI}). 
From the results, it can be seen that the introduction of cross-channel dependencies do not always enhance the prediction accuracy, and this was consistent across other datasets as well. Scaling rate $\beta$ can balance the emphasis on temporal dependencies and cross-channel dependencies in multivariate time series data and boost the model performance.

\begin{figure}[!t]
    \centering
    \includegraphics[width=0.41\textwidth]{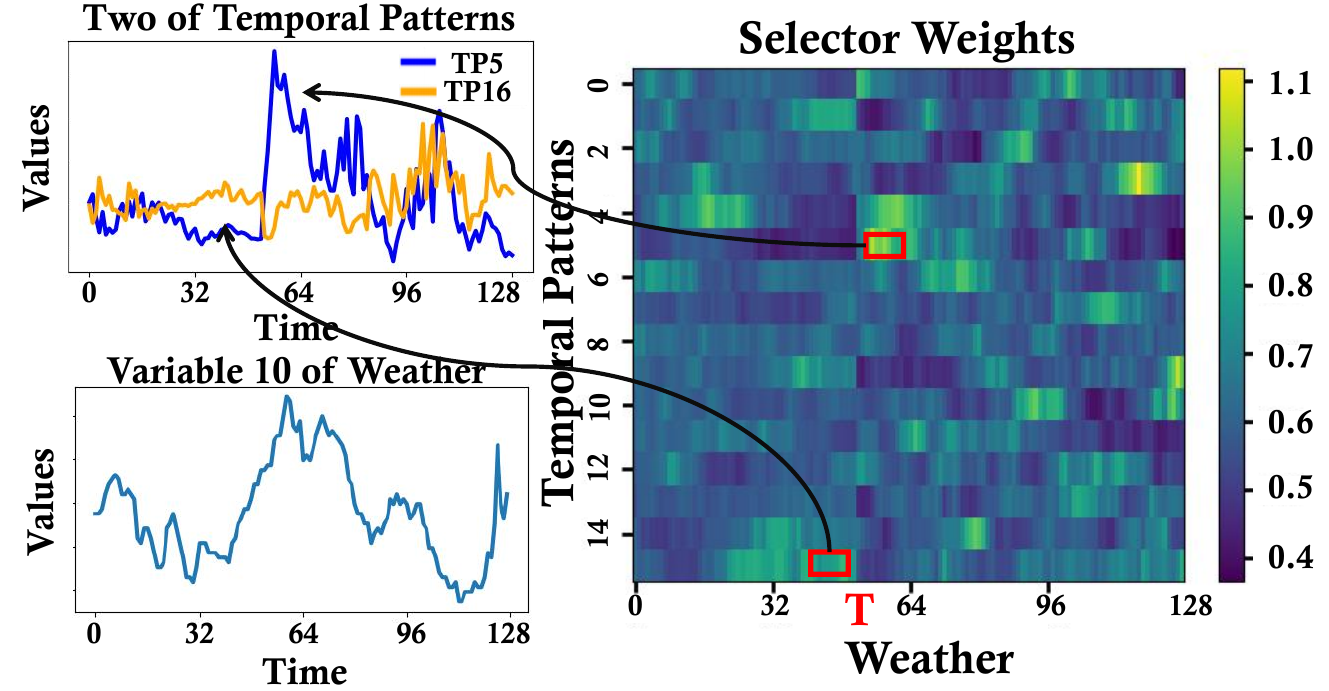} 
    \caption{\name guides the prediction by assigning greater weight to the dominant time pattern.
    }
    \label{fig:interpretability}
\end{figure}

For AMS, we demonstrate the performance improvement is not solely due to the enlarged model size, but rather the integration of temporal pattern information, we devised the following experiments:
To demonstrate the utilization of sequential information, we replaced the temporal pattern embedding with random orders as \textbf{RandomOrder}. 
To demonstrate the effectiveness of AMS aggregation, we replaced the TP-Selector with the average weighting as \textbf{AverageWeight}, treating different temporal patterns equally. 
We also remove AMS (\textbf{w/o AMS}) and replace it with a single linear predictor. 
\textbf{RandomOrder}, \textbf{AverageWeight} and \textbf{w/o AMS} all cause a performance decline. 
This illustrates that, compared to self-attention whose permutation-invariant nature results in the loss of sequential information, AMS inherently maintains the sequential order and makes better use of temporal relationships by identifying dominant patterns that change over time. Compared to simple averaging like timemixer~\cite{timemixer}, AMS adaptively assigns corresponding weights to different temporal patterns, resulting in more accurate predictions.

\paragraph{Model Interpretablilty.} 

To provide an intuitive understanding of AMS, we plotted selector weights in \cref{fig:interpretability}.
Compared to a single linear predictor, AMS offers stronger generalization capabilities and improved interpretability. 
Before and after the time step T, the temporal variations are respectively dominated by TP16 and TP5. Before T, the predicted data resembles the trend of TP16, both exhibiting a downward fluctuation. However, after T, the predicted data resembles TP5, which suddenly shows a significant increase. AMD recognizes this changing dominant role over time and adaptively assigns them higher weights.

\begin{figure}
    \centering
    \includegraphics[width=0.42\textwidth]{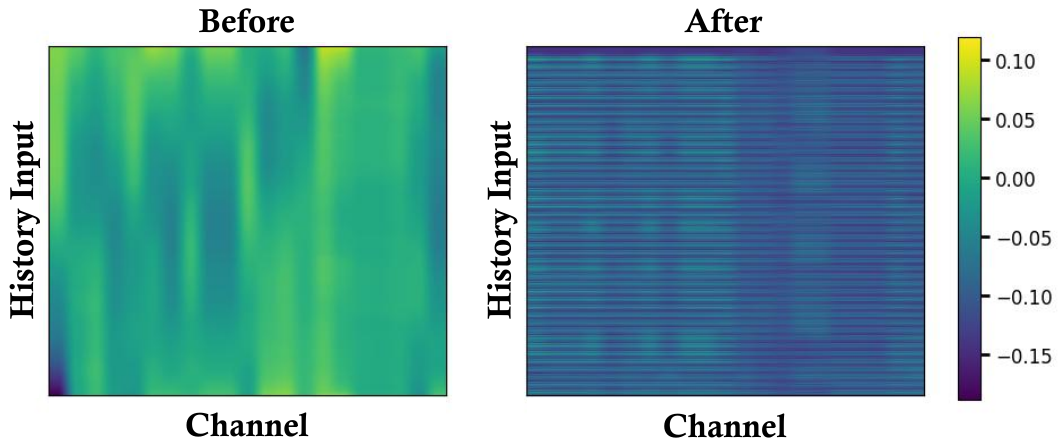} 
    \caption{Cross-channel dependencies may lead to deviations from the original distribution. 
    }
    \label{fig:crosschannel}
\end{figure}

To offer a clear conceptualization of the balance of emphasis on temporal dependencies and cross-channel dependencies mentioned above, we visualize the learned dependencies as shown in \cref{fig:crosschannel}. Compared to the temporal dependencies, especially when the target variable is not correlated with other covariates, cross-channel dependencies tend to smooth out the variability in the target variable, causing its distribution to deviate from what would be expected based solely on its own past values.

\begin{figure}
    \centering
    \includegraphics[width=0.4\textwidth]{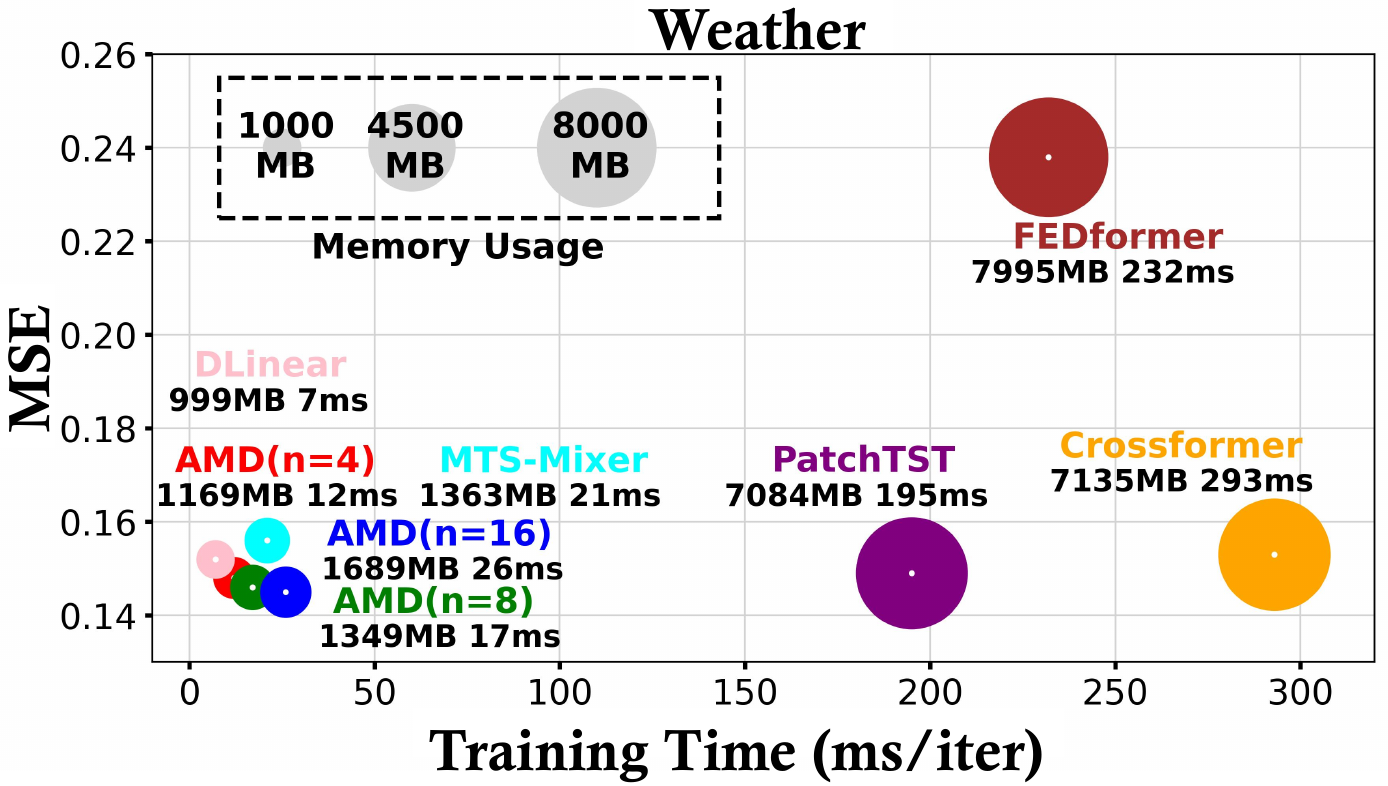} 
    \caption{Memory usage (MB), training time (ms/iter), and MSE comparisons on the Weather dataset. The input and predicted lengths are set to 512 and 96, respectively. 
    }
    \label{fig:efficiency}
\end{figure}

\paragraph{Dense MoE Strategy and Sparse MoE Strategy.}
In theory, the information contained in each temporal pattern is useful and discarding any one of them would result in loss of information.
Therefore, we adopt the dense MoE strategy, where dominant temporal patterns are given larger weights, while others are given smaller weights instead of being set to 0. We conduct ablation experiments as shown in \cref{tab:ablation}. Compared to Dense MoE, Sparse MoE shows increased prediction errors. This observation highlights the consequence of omitting non-dominant temporal pattern information, which invariably leads to a degradation in performance.

\paragraph{Efficiency Analysis.} 
We thoroughly compare the training time and memory usage of various baselines on the Weather dataset, using the official model configurations and the same batch size. The results, shown in \cref{fig:efficiency}, indicate that the efficiency of \name outperforms other Transformer-based models and MLP-based models with a relatively small number of parameters. Furthermore, as the number of predictors increases, the prediction error (MSE) gradually decreases, but the training time and memory usage increase significantly. To strike a balance, we set the number of predictors to 8 across all experiments.

\paragraph{Analysis on the balance of the TP-Selector.} 
We conduct experiments on the scaling rate of the load balancing loss, denoted by $\lambda_1$ in \cref{eq:loss}. As shown in \cref{tab:ablation}, utilizing $\mathcal{L}_\text{selector}$ results in significantly improved performance, exceeding $11.2\%$ in terms of MSE compared to when $\lambda_1=0.0$. This underscores the crucial role of implementing load-balancing losses.
Furthermore, the selector weights in the \cref{fig:interpretability} do not tend to favor specific temporal patterns, addressing the load imbalance issue in sparse MoE structures.

\renewcommand{\arraystretch}{0.5}
\begin{table}
\setlength{\tabcolsep}{3.5pt}
\scriptsize
\centering
\begin{threeparttable}
\begin{tabular}{c|c|cc|cc|cc|cc}
\toprule

\multicolumn{2}{c}{\scalebox{1.1}{Models}} & \multicolumn{2}{c}{DLinear} & \multicolumn{2}{c}{$+$ MDM \& AMS} & \multicolumn{2}{c}{MTS-Mixers} & \multicolumn{2}{c}{$+$ MDM \& AMS} \\

 \cmidrule(lr){3-4} \cmidrule(lr){5-6} \cmidrule(lr){7-8} \cmidrule(lr){9-10} 

\multicolumn{2}{c}{Metric} & MSE & MAE & MSE & MAE & MSE & MAE & MSE & MAE \\
 
\toprule

\multirow{4}{*}{\rotatebox{90}{Weather}}
& 96  & 0.152 & 0.237 & \best{0.146} & \best{0.212} & 0.156 & 0.206 & \best{0.155} & \best{0.203} \\
& 192 & 0.220 & 0.282 & \best{0.194} & \best{0.261} & \best{0.199} & \best{0.248} & 0.202 & 0.251 \\
& 336 & 0.265 & 0.319 & \best{0.245} & \best{0.305} & 0.249 & 0.291 & \best{0.244} & \best{0.286} \\
& 720 & 0.323 & 0.362 & \best{0.313} & \best{0.356} & 0.336 & 0.343 & \best{0.326} & \best{0.337} \\

\midrule

\multicolumn{2}{c|}{\textit{Imp.}} & - & - & 6.46$\%$ & 5.50$\%$ & - & - & 1.38$\%$ & 1.01$\%$ \\

\midrule

\multirow{4}{*}{\rotatebox{90}{ECL}}
& 96  & 0.153 & \best{0.237} & \best{0.150} & 0.244 & 0.141 & 0.243 & \best{0.137} & \best{0.239} \\
& 192 & \best{0.152} & \best{0.249} & 0.159 & 0.256 & 0.163 & 0.261 & \best{0.160} & \best{0.258} \\
& 336 & 0.169 & 0.267 & \best{0.167} & \best{0.265} & 0.176 & 0.277 & \best{0.170} & \best{0.271} \\
& 720 & 0.233 & 0.344 & \best{0.221} & \best{0.313} & 0.212 & 0.308 & \best{0.203} & \best{0.303} \\

\midrule

\multicolumn{2}{c|}{\textit{Imp.}} & - & - & 1.41$\%$ & 1.73$\%$ & - & - & 3.18$\%$ & $1.65\%$ \\

\bottomrule

\end{tabular}
\caption{Comparative impact of MDM \& AMS on different baselines. \textit{Imp.} represents the average percentage improvement of MDM \& AMS compared to the original methods.}
\label{tab:plugin}
\end{threeparttable}
\end{table}
\renewcommand{\arraystretch}{0.5}

\paragraph{Analysis on AMD as a plugin.} 
Finally, we explore whether our proposed MoE-based method can yield an improvement in the performance of other TSF methods. We selected DLinear~\cite{dlinear} and MTS-Mixers~\cite{mtsmixers} as the baselines. After integrating the MDM and AMS modules, the predictive capabilities of all these models are enhanced as shown in \cref{tab:plugin}, while maintaining the same computational resource requirements.

\section{Conclusion}
In this paper, we propose the Adaptive Multi-Scale Decomposition (\name) framework for time series forecasting to address the inherent complexity of time series data by decomposing it into multiple temporal patterns at various scales and adaptively aggregating these patterns. Comprehensive experiments demonstrate that \name consistently achieves state-of-the-art performance in both long-term and short-term forecasting tasks across various datasets, showcasing superior efficiency and effectiveness. 

\section{Acknowledgements}
This work is supported  by the National Key R\&D Program of China (Grant no. 2022YFB4501704), the National Science Fundation of China (Grant no. 62472317), the Fundamental Research Funds for the Central Universities, the Shanghai Science and Technology Innovation Action Plan Project (Grant no. 22YS1400600 and 24692118300). 
This work is also supported in part by the National Natural Science Foundation of China, under Grant (62302309,),  Shenzhen Science and Technology Program (JCYJ20220818101014030).

\bibliography{aaai25}

\begin{thebibliography}{45}
\providecommand{\natexlab}[1]{#1}

\bibitem[{Chen et~al.(2023)Chen, Li, Arik, Yoder, and Pfister}]{googletsmixer}
Chen, S.-A.; Li, C.-L.; Arik, S.~O.; Yoder, N.~C.; and Pfister, T. 2023.
\newblock {TSM}ixer: An All-{MLP} Architecture for Time Series Forecast-ing.
\newblock \emph{Transactions on Machine Learning Research}.

\bibitem[{Cheng et~al.(2018)Cheng, Liu, Niu, and Zhang}]{8425030}
Cheng, D.; Liu, Y.; Niu, Z.; and Zhang, L. 2018.
\newblock Modeling Similarities Among Multi-Dimensional Financial Time Series.
\newblock \emph{IEEE Access}, 6: 43404--43413.

\bibitem[{Cheng et~al.(2022)Cheng, Yang, Xiang, and Liu}]{fints}
Cheng, D.; Yang, F.; Xiang, S.; and Liu, J. 2022.
\newblock Financial time series forecasting with multi-modality graph neural network.
\newblock \emph{Pattern Recognition}, 121: 108218.

\bibitem[{Dai et~al.(2024)Dai, Wu, Liu, Li, Bao, Jiang, and Xia}]{dai2024periodicity}
Dai, T.; Wu, B.; Liu, P.; Li, N.; Bao, J.; Jiang, Y.; and Xia, S.-T. 2024.
\newblock Periodicity Decoupling Framework for Long-term Series Forecasting.
\newblock In \emph{International Conference on Learning Representations}.

\bibitem[{Das et~al.(2023)Das, Kong, Leach, Mathur, Sen, and Yu}]{tide}
Das, A.; Kong, W.; Leach, A.; Mathur, S.~K.; Sen, R.; and Yu, R. 2023.
\newblock Long-term Forecasting with {T}i{DE}: Time-series Dense Encoder.
\newblock \emph{Transactions on Machine Learning Research}.

\bibitem[{Han, Ye, and Zhan(2023)}]{civscd}
Han, L.; Ye, H.-J.; and Zhan, D.-C. 2023.
\newblock The Capacity and Robustness Trade-off: Revisiting the Channel Independent Strategy for Multivariate Time Series Forecasting.
\newblock \emph{arXiv preprint arXiv:2304.05206}.

\bibitem[{Holt(2004)}]{HOLT}
Holt, C.~C. 2004.
\newblock Forecasting seasonals and trends by exponentially weighted moving averages.
\newblock \emph{International Journal of Forecasting}, 20(1): 5--10.

\bibitem[{Jacobs et~al.(1991)Jacobs, Jordan, Nowlan, and Hinton}]{moe}
Jacobs, R.~A.; Jordan, M.~I.; Nowlan, S.~J.; and Hinton, G.~E. 1991.
\newblock {Adaptive Mixtures of Local Experts}.
\newblock \emph{Neural Computation}, 3(1): 79--87.

\bibitem[{Jia et~al.(2023)Jia, Lin, Hao, Lin, Guo, and Wan}]{WITRANWI}
Jia, Y.; Lin, Y.; Hao, X.; Lin, Y.; Guo, S.; and Wan, H. 2023.
\newblock W{ITRAN}: Water-wave Information Transmission and Recurrent Acceleration Network for Long-range Time Series Forecasting.
\newblock In \emph{Advances in Neural Information Processing Systems}.

\bibitem[{Jordan(1994)}]{moe1}
Jordan, M. 1994.
\newblock Hierarchical mixtures of experts and the EM algorithm.
\newblock In \emph{IEE Colloquium on Advances in Neural Networks for Control and Systems}, 1/1--1/3.

\bibitem[{Kim et~al.(2022{\natexlab{a}})Kim, Cho, Kim, Park, and Choo}]{traffic}
Kim, D.; Cho, Y.; Kim, D.; Park, C.; and Choo, J. 2022{\natexlab{a}}.
\newblock Residual Correction in Real-Time Traffic Forecasting.
\newblock In \emph{Proceedings of the ACM International Conference on Information \& Knowledge Management}, 962–971. New York, NY, USA: Association for Computing Machinery.
\newblock ISBN 9781450392365.

\bibitem[{Kim et~al.(2022{\natexlab{b}})Kim, Kim, Tae, Park, Choi, and Choo}]{revin}
Kim, T.; Kim, J.; Tae, Y.; Park, C.; Choi, J.; and Choo, J. 2022{\natexlab{b}}.
\newblock Reversible Instance Normalization for Accurate Time-Series Forecasting against Distribution Shift.
\newblock In \emph{International Conference on Learning Representations}.

\bibitem[{Kim, Lim, and Han(2024)}]{kim2024scaling}
Kim, Y.; Lim, H.; and Han, D. 2024.
\newblock Scaling Beyond the {GPU} Memory Limit for Large Mixture-of-Experts Model Training.
\newblock In \emph{International Conference on Machine Learning}.

\bibitem[{Li et~al.(2023{\natexlab{a}})Li, Shen, Yang, Wang, Ren, Che, Zhang, and Liu}]{sparse}
Li, B.; Shen, Y.; Yang, J.; Wang, Y.; Ren, J.; Che, T.; Zhang, J.; and Liu, Z. 2023{\natexlab{a}}.
\newblock Sparse Mixture-of-Experts are Domain Generalizable Learners.
\newblock In \emph{International Conference on Learning Representations}.

\bibitem[{Li et~al.(2024)Li, Qi, Li, and Xu}]{rlinear}
Li, Z.; Qi, S.; Li, Y.; and Xu, Z. 2024.
\newblock Revisiting Long-term Time Series Forecasting: An Investigation on Affine Mapping.

\bibitem[{Li et~al.(2023{\natexlab{b}})Li, Rao, Pan, and Xu}]{mtsmixers}
Li, Z.; Rao, Z.; Pan, L.; and Xu, Z. 2023{\natexlab{b}}.
\newblock M{TS-M}ixers: Multivariate Time Series Forecasting via Factorized Temporal and Channel Mixing.
\newblock \emph{arXiv preprint arXiv:2302.04501}.

\bibitem[{Lin et~al.(2023)Lin, Lin, Wu, Zhao, Mo, and Zhang}]{segrnn}
Lin, S.; Lin, W.; Wu, W.; Zhao, F.; Mo, R.; and Zhang, H. 2023.
\newblock Seg{RNN}: Segment Recurrent Neural Network for Long-Term Time Series Forecasting.
\newblock \emph{arXiv preprint arXiv:2308.11200}.

\bibitem[{LIU et~al.(2022)LIU, Zeng, Chen, Xu, LAI, Ma, and Xu}]{scinet}
LIU, M.; Zeng, A.; Chen, M.; Xu, Z.; LAI, Q.; Ma, L.; and Xu, Q. 2022.
\newblock SCINet: Time Series Modeling and Forecasting with Sample Convolution and Interaction.
\newblock In Koyejo, S.; Mohamed, S.; Agarwal, A.; Belgrave, D.; Cho, K.; and Oh, A., eds., \emph{Advances in Neural Information Processing Systems}, volume~35, 5816--5828. Curran Associates, Inc.

\bibitem[{Liu et~al.(2022)Liu, Yu, Liao, Li, Lin, Liu, and Dustdar}]{pyraformer}
Liu, S.; Yu, H.; Liao, C.; Li, J.; Lin, W.; Liu, A.~X.; and Dustdar, S. 2022.
\newblock Pyraformer: Low-Complexity Pyramidal Attention for Long-Range Time Series Modeling and Forecasting.
\newblock In \emph{International Conference on Learning Representations}.

\bibitem[{Liu et~al.(2024{\natexlab{a}})Liu, Hu, Zhang, Wu, Wang, Ma, and Long}]{itransformer}
Liu, Y.; Hu, T.; Zhang, H.; Wu, H.; Wang, S.; Ma, L.; and Long, M. 2024{\natexlab{a}}.
\newblock i{T}ransformer: Inverted Transformers Are Effective for Time Series Forecasting.
\newblock In \emph{International Conference on Learning Representations}.

\bibitem[{Liu et~al.(2024{\natexlab{b}})Liu, Wang, Vaidya, Ruehle, Halverson, Soljačić, Hou, and Tegmark}]{kan}
Liu, Z.; Wang, Y.; Vaidya, S.; Ruehle, F.; Halverson, J.; Soljačić, M.; Hou, T.~Y.; and Tegmark, M. 2024{\natexlab{b}}.
\newblock KAN: Kolmogorov-Arnold Networks.
\newblock \emph{arXiv preprint arXiv:2404.19756}.

\bibitem[{Ni et~al.(2024)Ni, Lin, Wang, and Fanti}]{linearexperts}
Ni, R.; Lin, Z.; Wang, S.; and Fanti, G. 2024.
\newblock Mixture-of-Linear-Experts for Long-term Time Series Forecasting.
\newblock \emph{arXiv preprint arXiv:2312.06786}.

\bibitem[{Nie et~al.(2023)Nie, Nguyen, Sinthong, and Kalagnanam}]{patchtst}
Nie, Y.; Nguyen, N.~H.; Sinthong, P.; and Kalagnanam, J. 2023.
\newblock A Time Series is Worth 64 Words: Long-term Forecasting with Transformers.
\newblock In \emph{International Conference on Learning Representations}.

\bibitem[{Oreshkin et~al.(2020)Oreshkin, Carpov, Chapados, and Bengio}]{nbeats}
Oreshkin, B.~N.; Carpov, D.; Chapados, N.; and Bengio, Y. 2020.
\newblock N{-BEATS}: Neural basis expansion analysis for interpretable time series forecasting.
\newblock In \emph{International Conference on Learning Representations}.

\bibitem[{Qiu et~al.(2024)Qiu, Wu, Lin, Guo, Hu, and Yang}]{duet}
Qiu, X.; Wu, X.; Lin, Y.; Guo, C.; Hu, J.; and Yang, B. 2024.
\newblock DUET: Dual Clustering Enhanced Multivariate Time Series Forecasting.
\newblock \emph{arXiv preprint arXiv:2412.10859}.

\bibitem[{Shabani et~al.(2023)Shabani, Abdi, Meng, and Sylvain}]{scaleformer}
Shabani, M.~A.; Abdi, A.~H.; Meng, L.; and Sylvain, T. 2023.
\newblock Scaleformer: Iterative Multi-scale Refining Transformers for Time Series Forecasting.
\newblock In \emph{International Conference on Learning Representations}.

\bibitem[{Shazeer et~al.(2017)Shazeer, Mirhoseini, Maziarz, Davis, Le, Hinton, and Dean}]{llm}
Shazeer, N.; Mirhoseini, A.; Maziarz, K.; Davis, A.; Le, Q.; Hinton, G.; and Dean, J. 2017.
\newblock Outrageously Large Neural Networks: The Sparsely-Gated Mixture-of-Experts Layer.
\newblock In \emph{International Conference on Learning Representations}.

\bibitem[{Tolstikhin et~al.(2021)Tolstikhin, Houlsby, Kolesnikov, Beyer, and Zhai}]{googlemlpmixer}
Tolstikhin, I.~O.; Houlsby, N.; Kolesnikov, A.; Beyer, L.; and Zhai. 2021.
\newblock M{LP-M}ixer: An all-{MLP} Architecture for Vision.
\newblock In Ranzato, M.; Beygelzimer, A.; Dauphin, Y.; Liang, P.; and Vaughan, J.~W., eds., \emph{Advances in Neural Information Processing Systems}, volume~34, 24261--24272. Curran Associates, Inc.

\bibitem[{Vaswani et~al.(2017)Vaswani, Shazeer, Parmar, Uszkoreit, Jones, Gomez, Kaiser, and Polosukhin}]{vaswani2017attention}
Vaswani, A.; Shazeer, N.; Parmar, N.; Uszkoreit, J.; Jones, L.; Gomez, A.~N.; Kaiser, {\L}.; and Polosukhin, I. 2017.
\newblock Attention is all you need.
\newblock \emph{Advances in Neural Information Processing Systems}, 30.

\bibitem[{Volkovs, Urtans, and Caune(2024)}]{weather}
Volkovs, K.; Urtans, E.; and Caune, V. 2024.
\newblock Primed UNet-LSTM for Weather Forecasting.
\newblock In \emph{Proceedings of the International Conference on Advances in Artificial Intelligence}, ICAAI '23, 13–17. New York, NY, USA: Association for Computing Machinery.
\newblock ISBN 9798400708985.

\bibitem[{Wang et~al.(2023)Wang, Peng, Huang, Wang, Chen, and Xiao}]{micn}
Wang, H.; Peng, J.; Huang, F.; Wang, J.; Chen, J.; and Xiao, Y. 2023.
\newblock M{ICN}: Multi-scale Local and Global Context Modeling for Long-term Series Forecasting.
\newblock In \emph{International Conference on Learning Representations}.

\bibitem[{Wang et~al.(2024)Wang, Wu, Shi, Hu, Luo, Ma, Zhang, and ZHOU}]{timemixer}
Wang, S.; Wu, H.; Shi, X.; Hu, T.; Luo, H.; Ma, L.; Zhang, J.~Y.; and ZHOU, J. 2024.
\newblock Time{M}ixer: Decomposable Multiscale Mixing for Time Series Forecasting.
\newblock In \emph{International Conference on Learning Representations}.

\bibitem[{Wu et~al.(2023)Wu, Hu, Liu, Zhou, Wang, and Long}]{timesnet}
Wu, H.; Hu, T.; Liu, Y.; Zhou, H.; Wang, J.; and Long, M. 2023.
\newblock Times{N}et: Temporal 2D-Variation Modeling for General Time Series Analysis.
\newblock In \emph{International Conference on Learning Representations}.

\bibitem[{Wu et~al.(2021)Wu, Xu, Wang, and Long}]{autoformer}
Wu, H.; Xu, J.; Wang, J.; and Long, M. 2021.
\newblock Autoformer: Decomposition Transformers with Auto-Correlation for Long-Term Series Forecasting.
\newblock In Ranzato, M.; Beygelzimer, A.; Dauphin, Y.; Liang, P.; and Vaughan, J.~W., eds., \emph{Advances in Neural Information Processing Systems}, volume~34, 22419--22430. Curran Associates, Inc.

\bibitem[{Wu et~al.(2020)Wu, Pan, Long, Jiang, Chang, and Zhang}]{mtgnn}
Wu, Z.; Pan, S.; Long, G.; Jiang, J.; Chang, X.; and Zhang, C. 2020.
\newblock Connecting the {D}ots: {M}ultivariate {T}ime {S}eries {F}orecasting with {G}raph {N}eural {N}etworks.
\newblock In \emph{Proceedings of the ACM SIGKDD International Conference on Knowledge Discovery \& Data Mining}, KDD '20, 753–763. New York, NY, USA: Association for Computing Machinery.
\newblock ISBN 9781450379984.

\bibitem[{Xu, Zeng, and Xu(2024)}]{fits}
Xu, Z.; Zeng, A.; and Xu, Q. 2024.
\newblock {FITS}: Modeling Time Series with \$10k\$ Parameters.
\newblock In \emph{International Conference on Learning Representations}.

\bibitem[{Xue et~al.(2024)Xue, Zheng, Fu, Ni, Zheng, Zhou, and You}]{openmoe}
Xue, F.; Zheng, Z.; Fu, Y.; Ni, J.; Zheng, Z.; Zhou, W.; and You, Y. 2024.
\newblock OpenMoE: An Early Effort on Open Mixture-of-Experts Language Models.
\newblock In \emph{International Conference on Machine Learning}.

\bibitem[{Xue and Salim(2023)}]{energy}
Xue, H.; and Salim, F.~D. 2023.
\newblock Utilizing Language Models for Energy Load Forecasting.
\newblock In \emph{Proceedings of the ACM International Conference on Systems for Energy-Efficient Buildings, Cities, and Transportation}, BuildSys '23, 224–227. New York, NY, USA: Association for Computing Machinery.
\newblock ISBN 9798400702303.

\bibitem[{Yi et~al.(2023)Yi, Zhang, Fan, He, Hu, Wang, An, Cao, and Niu}]{fouriergnn}
Yi, K.; Zhang, Q.; Fan, W.; He, H.; Hu, L.; Wang, P.; An, N.; Cao, L.; and Niu, Z. 2023.
\newblock Fourier{GNN}: Rethinking Multivariate Time Series Forecasting from a Pure Graph Perspective.
\newblock In Oh, A.; Naumann, T.; Globerson, A.; Saenko, K.; Hardt, M.; and Levine, S., eds., \emph{Advances in Neural Information Processing Systems}, volume~36, 69638--69660. Curran Associates, Inc.

\bibitem[{Zeng et~al.(2023)Zeng, Chen, Zhang, and Xu}]{dlinear}
Zeng, A.; Chen, M.; Zhang, L.; and Xu, Q. 2023.
\newblock Are Transformers Effective for Time Series Forecasting?
\newblock \emph{Proceedings of the AAAI Conference on Artificial Intelligence}, 37(9): 11121--11128.

\bibitem[{Zhang et~al.(2022)Zhang, Zhang, Cao, Bian, Yi, Zheng, and Li}]{lightts}
Zhang, T.; Zhang, Y.; Cao, W.; Bian, J.; Yi, X.; Zheng, S.; and Li, J. 2022.
\newblock Less Is More: Fast Multivariate Time Series Forecasting with Light Sampling-oriented MLP Structures.
\newblock \emph{arXiv preprint arXiv:2207.01186}.

\bibitem[{Zhang and Yan(2023)}]{crossformer}
Zhang, Y.; and Yan, J. 2023.
\newblock Crossformer: Transformer Utilizing Cross-Dimension Dependency for Multivariate Time Series Forecasting.
\newblock In \emph{International Conference on Learning Representations}.

\bibitem[{Zhou et~al.(2022{\natexlab{a}})Zhou, MA, Wang, Wen, Sun, Yao, Yin, and Jin}]{film}
Zhou, T.; MA, Z.; Wang, X.; Wen, Q.; Sun, L.; Yao, T.; Yin, W.; and Jin, R. 2022{\natexlab{a}}.
\newblock Fi{LM}: Frequency improved Legendre Memory Model for Long-term Time Series Forecasting.
\newblock In Koyejo, S.; Mohamed, S.; Agarwal, A.; Belgrave, D.; Cho, K.; and Oh, A., eds., \emph{Advances in Neural Information Processing Systems}, volume~35, 12677--12690. Curran Associates, Inc.

\bibitem[{Zhou et~al.(2022{\natexlab{b}})Zhou, Ma, Wen, Wang, Sun, and Jin}]{fedformer}
Zhou, T.; Ma, Z.; Wen, Q.; Wang, X.; Sun, L.; and Jin, R. 2022{\natexlab{b}}.
\newblock F{ED}former: Frequency enhanced decomposed transformer for long-term series forecasting.
\newblock In \emph{International Conference on Machine Learning}, 27268--27286. PMLR.

\bibitem[{Zhu et~al.(2024)Zhu, Li, Hu, Liu, Cheng, and Liang}]{lsrigru}
Zhu, P.; Li, Y.; Hu, Y.; Liu, Q.; Cheng, D.; and Liang, Y. 2024.
\newblock LSR-IGRU: Stock Trend Prediction Based on Long Short-Term Relationships and Improved GRU.
\newblock In \emph{Proceedings of the 33rd ACM International Conference on Information and Knowledge Management}, CIKM '24, 5135–5142. New York, NY, USA: Association for Computing Machinery.
\newblock ISBN 9798400704369.

\end{thebibliography}

\appendix
\section{Proof of Theorem 1}\label{sec:property-proof}
\setcounter{theorem}{1}
\addtocounter{theorem}{-1}
\begin{theorem}
Let multi-scale mixing representation $g(x)$, where $g(x)\in\mathbb{R}^{1\times L}$ (for simplicity, we consider univariate sequences) and the original sequence $f(x)$ is Lipschitz smooth with constant $\mathcal{K}$ ($i.e.\ \ \left| \frac{f(a)-f(b)}{a-b} \right|\leq\mathcal{K}$), then there exists a linear model such that $\left|y_t-\hat{y_t}\right|$ is bounded, $\forall t=1,...,T$.
\end{theorem}
\begin{proof}
We first prove that the downsampled sequence $f_i(x), \forall i=1,...,n$ possesses Lipschitz smooth. $\forall t\in\mathcal{D}(f_i(x)), t\in\mathcal{D}(f(x))$, where $\mathcal{D}$ means the domain of the sequence. 
Therefore, we can conclude that, 
\begin{align}
&\left| \frac{f_1(a)-f_1(b)}{a - b} \right| \\
=&\frac{1}{d} \left| \frac{\sum_{j=ad-d+1}^{ad}f(j) - \sum_{j=bd-d+1}^{bd}f(j)}{a - b} \right|\\
=&\frac{1}{d} \left| \frac{\sum_{j=0}^{d-1}[f(ad+j-d+1)-f(bd+j-d+1)]}{ad - bd} \right|\\
\leq& \frac{1}{d} \cdot d\mathcal{K}\\
\leq& \mathcal{K}
\end{align}
Similarly, by mathematical induction in a bottom-up manner, we can prove that $f_i(x), \forall i=1,...,n$ possesses Lipschitz smooth.

Then, we prove multi-scale mixing representation $g_i(x), \forall i=0,...,n$ possesses Lipschitz smooth. According to the property of linear combination, we have $g_i(t) = f_i(t) + \sum_{j=0}^{\big[\frac{n}{d^{i+1}}\big]}g_{i+1}(t)W_{i}(j, t)$, where $W_i(j, t)$ represents the $j$-th row and $t$-th column. So we have,

\begin{align}
&\left| \frac{g_{n-1}(a)-g_{n-1}(b)}{a - b} \right| \\
=&\left| \frac{1}{a - b}\right| \bigg|f_{n-1}(a)-f_{n-1}(b) + \\
&\sum_{j=0}^{\big[\frac{L}{d^n}\big]}g_{n}(a)W_{n-1}(j, a) - \sum_{j=0}^{\big[\frac{L}{d^{i+1}}\big]}g_{n}(b)W_{n-1}(j, b)  \bigg|\\ 
\leq&\left| \frac{f_{n-1}(a)-f_{n-1}(b)}{a - b} \right| + \\
&\left| \frac{\sum_{j=0}^{\big[\frac{L}{d^n}\big]}g_{n}(a)W_{n-1}(j, a) - \sum_{j=0}^{\big[\frac{L}{d^{n}}\big]}g_{n}(b)W_{n-1}(j, b)}{a - b} \right|\\
\leq&\mathcal{K} + \left| \sum_{j=0}^{\big[\frac{L}{d^{n}}\big]}\frac{g_{n}(a)-g_{n}(b)}{a - b} \cdot max\{W_{n-1}(j, a), W_{n-1}(j, b)\}\right|\\
=&\mathcal{K} + \left| \sum_{j=0}^{\big[\frac{L}{d^{n}}\big]}\frac{f_{n}(a)-f_{n}(b)}{a - b} \cdot max\{W_{n-1}(j, a), W_{n-1}(j, b)\}\right|\\
\leq&\mathcal{K}\cdot\left(1 + \left| \sum_{j=0}^{\big[\frac{L}{d^{n}}\big]} max\{W_{n-1}(j, a), W_{n-1}(j, b)\}\right|\right)
\end{align}
Therefore, \(g_{n-1}\) possesses smooth due to \(W_i\) being a constant matrix. Similarly, by mathematical induction in a top-down manner, we can prove that $g_i(x), \forall i=0,...,n$ possesses Lipschitz smooth.

Subsequently, we prove $\left|y_t^{m}-\hat{y_t^{m}}\right|$ is bounded, Where $y_t^m$ means the TP mixed observed data, and $\hat{y_t^m}$ represents the predicted TP mixed data. We set the period of the finest granularity time pattern as P. If there is no periodicity, then P tends to positive infinity ($P\xrightarrow{}+\infty$).
\begin{align}
    y_t^{m}&=g(L + t)=g(P + 1 + t)\\
    \hat{y_t^{m}}&=g(L + t)A\oplus b
\end{align}
Let $A\in\mathbb{R}^{L\times T}$, and
\begin{align}
A_{tj}=
\begin{cases} 
1,  & \text{if }j = P + 1 \text{ or }j = (t \% P) + 1 \\
-1, & \text{if }j = 1 \\
0, & \text{otherwise}
\end{cases}
\ \ \ \ ,\ \ \ \ b_t=0
\end{align}
Then,
\begin{align}
\hat{y_t^{m}}=g(t\%P+1)-g(1)+g(P+1)
\end{align}
So,
\begin{align}
&\left| y_t^{m}-\hat{y_t^{m}} \right|\\
=&\left| [g(P+1+t)-g(P+1)] - [g(t\%P+1)-g(1)] \right|\\
\leq&\left|g(P+1+t)-g(P+1)\right|+\left|g(t\%P+1)-g(1)\right|\\
\leq&\mathcal{K}(t+t\%P)
\end{align}

Finally, we employ a weighted pattern predictor on the TP mixed data. Since we solely apply internal averaging operations during this process, the resulting remains bounded. This is due to the inherent property of internal averaging to smooth out fluctuations within the data without introducing significant variations beyond certain bounds. Therefore, $\left|y_t-\hat{y_t}\right|$ is bounded.
\end{proof}

\begin{algorithm}[b!]
\caption{The Overall Architecture of AMD}\label{algorithm}
\begin{algorithmic}
    \State \textbf{Input:} look-back sequence $X\in\mathbb{R}^{L\times C}$.
    \State \textbf{Parameter:} DDI block number $n$.
    \State \textbf{Output:} Predictions $\mathbf{\hat{Y}}$.
    \State{$\mathbf{X}=\text{RevIn}(\mathbf{X},\ \ \text{'norm'})$}
    \State{$\mathbf{X}=\mathbf{X^T}$}\Comment{$\mathbf{X}\in\mathbb{R}^{C\times L}$}
    \State{$\mathbf{U}=\text{MDM}(\mathbf{X})$}\Comment{$\mathbf{U}\in\mathbb{R}^{C\times L}$}
    \For{$i$ \text{in} $\{1,...,n\}$}
        \State{$\mathbf{U}=\text{LayerNorm}(\mathbf{U})$}
        \State{$\mathbf{V}_0^l=\mathbf{U}_0^P$}
        \While{$j \leq L$}
        \State{$\mathbf{Z}_{j}^{j+P} = \mathbf{U}_{j}^{j+P} + \text{FeedForward}(\mathbf{V}_{j-P}^{j})$}
        \State{$\mathbf{V}_{j}^{j+P} = \mathbf{Z}_{j}^{j+P} + \beta\cdot \text{FeedForward}((\mathbf{Z}_{j}^{j+P})^T)^T$}
        \State{$j=j+P$}
        \EndWhile
    \EndFor
    \State{$\mathbf{v}=\text{Split}(\mathbf{V})$}
    \For{$i$ \text{in} $\{1,...,C\}$}
        \State{$\mathbf{S}=\text{TP-Selector}(\mathbf{u})$}\Comment{$\mathbf{S}\in\mathbb{R}^{m\times T}$}
        \State{$\mathbf{\hat{y}}=\text{SUM}(\mathbf{S}\odot \text{Predictor}(\mathbf{v}), dim=0)$} \Comment{$\mathbf{\hat{y}}\in\mathbb{R}^{1\times T}$}
    \EndFor
    \State{$\mathbf{\hat{Y}}=\mathbf{\hat{Y}^T}$}\Comment{$\mathbf{\hat{Y}}\in\mathbb{R}^{L\times C}$}
    \State{$\mathbf{\hat{Y}}=\text{RevIn}(\mathbf{\hat{Y}},\ \ \text{'denorm'})$}
    \State{Return $\mathbf{\hat{Y}}$}\Comment{Prediction Results.}
\end{algorithmic}
\end{algorithm}

\section{Detailed Algorithm Description}\label{sec:detailed algorithm descriptions}
The pseudocode of the AMD algorithm is shown in Algorithm 1. The algorithm initializes input data and parameters and perform normalization. The data is then iterated through MDM block to extract multi-scale information. DDI blocks subsequently performs aggregation operations. Following this, for each feature, TPSelector determines predictor weights, with outputs concatenated and weighted for prediction. The algorithm transposes and de-normalizes predictions before returning them.

\section{Details of Experiments}\label{sec:detailed experiment}
\subsection{Detailed Dataset Descriptions}\label{sec:detailed dataset descriptions}
Detailed dataset descriptions are shown in \cref{tab:detaildataset}. Dim denotes the number of channels in each dataset. Dataset Size denotes the total number of time points in (Train, Validation, Test) split respectively. Prediction Length denotes the future time points to be predicted and four prediction settings are included in each dataset. Frequency denotes the sampling interval of time points. Information refers to the meaning of the data.

\renewcommand{\arraystretch}{1.0}
\begin{table*}[htb]
\setlength{\tabcolsep}{7pt}
\tiny
\centering
\begin{threeparttable}
\caption{Detailed dataset descriptions.}
\begin{tabular}{c|c|c|c|c|c}
\toprule

Dataset & Dim & Prediction Length  & Dataset Size & Frequency & Information \\

\midrule
ETTh1, ETTh2 & 7 & \{96,192,336,720\} & (8545,2881,2881) & Hourly & Electricity \\
\midrule
ETTm1, ETTm2 & 7 & \{96,192,336,720\} & (34465,11521,11521) & 15min & Electricity \\
\midrule
Exchange & 8 & \{96,192,336,720\} & (5120,665,1422) & Daily & Economy \\
\midrule
Weather & 21 & \{96,192,336,720\} & (36792,5271,10540) & 10min & Weather \\
\midrule
ECL & 321 & \{96,192,336,720\} & (18317,2633,5261) & Hourly & Electricity \\
\midrule
Traffic & 862 & \{96,192,336,720\} & (12185,1757,3509) & Hourly & Transportation \\
\midrule
Solar-Energy & 137 & \{96,192,336,720\} & (36601,5161,10417) & 10min & Energy \\
\midrule
PEMS03 & 358 & 12 & (15617,5135,5135) & 5min & Transportation \\
\midrule
PEMS04 & 307 & 12 & (10172,3375,3375) & 5min & Transportation \\
\midrule
PEMS07 & 883 & 12 & (16911,5622,5622) & 5min & Transportation \\
\midrule
PEMS08 & 170 & 12 & (10690,3548,3548) & 5min & Transportation \\

\bottomrule

\end{tabular}

\label{tab:detaildataset}
\end{threeparttable}
\end{table*}
\renewcommand{\arraystretch}{1}

\subsection{Baseline Models}\label{sec:baseline models}
We briefly describe the selected baselines:

(1) MTS-Mixers~\cite{mtsmixers} is an MLP-based model utilizing two factorized modules to model the mapping between the input and the prediction sequence. The source code is available at \url{https://github.com/plumprc/MTS-Mixers}. 

(2) DLinear~\cite{dlinear} is an MLP-based model with just one linear layer, which unexpectedly outperforms Transformer-based models in long-term TSF. The source code is available at \url{https://github.com/cure-lab/LTSF-Linear}.

(3) TiDE~\cite{tide} is a simple and effective MLP-based encoder-decoder model. The source code is available at \url{https://github.com/thuml/Time-Series-Library}.

(4) PatchTST~\cite{patchtst} is a Transformer-based model utilizing patching and CI technique. It also enable effective pre-training and transfer learning across datasets. The source code is available at \url{https://github.com/yuqinie98/PatchTST}.

(5) iTransformer~\cite{itransformer} embeds each time series as variate tokens and is a fundamental backbone for time series forecasting. The source code is available at \url{https://github.com/thuml/iTransformer}.

(6) Crossformer~\cite{crossformer} is a Transformer-based model introducing the Dimension-Segment-Wise (DSW) embedding and Two-Stage Attention (TSA) layer to effectively capture cross-time and cross-dimension dependencies. The source code is available at \url{https://github.com/Thinklab-SJTU/Crossformer}.

(7) FEDformer~\cite{fedformer} is a Transformer-based model proposing seasonal-trend decomposition and exploiting the sparsity of time series in the frequency domain. The source code is available at \url{https://github.com/DAMO-DI-ML/ICML2022-FEDformer}.

(8) TimesNet~\cite{timesnet} is a CNN-based model with TimesBlock as a task-general
backbone. It transforms 1D time series into 2D tensors to capture intraperiod and interperiod variations. The source code is available at \url{https://github.com/thuml/TimesNet}.

(9) MICN~\cite{micn} is a CNN-based model combining local features and global correlations to capture the overall view of time series. The source code is available at \url{https://github.com/wanghq21/MICN}.

\subsection{Metric Details}
Regarding metrics, we utilize the mean square error (MSE) and mean absolute
error (MAE) for long-term forecasting. The calculations of these metrics are:
$$
MSE=\frac{1}{T}\sum_{0}^{T}(\hat{y_i}-y_i)^2
$$
$$
MAE=\frac{1}{T}\sum_{0}^{T}|\hat{y_i}-y_i|
$$

\renewcommand{\arraystretch}{0.95}
\begin{table}[ht]
\setlength{\tabcolsep}{3pt}
\scriptsize
\centering
\begin{threeparttable}
\caption{The hyper-parameters for different experimental settings.}
\begin{tabular}{c|c|c|c|c|c|c|c}
\toprule

Dataset & $P$ & $\alpha$ & Batch Size & Epochs & $n$ & Learning Rate & Layer Norm \\
\midrule
ETTh1 & 16 & 0.0 & 128 & 10 & 1 & $5\times e^{-5}$ & True \\
\midrule
ETTh2 & 4 & 1.0 & 128 & 10 & 1 & $5\times e^{-5}$ & False \\
\midrule
ETTm1 & 16 & 0.0 & 128 & 10 & 1 & $3\times e^{-5}$ & True \\
\midrule
ETTm2 & 8 & 0.0 & 128 & 10 & 1 & $1\times e^{-5}$ & True \\
\midrule
Exchange & 4 & 0.0 & 512 & 10 & 1 & $3\times e^{-4}$ & True \\
\midrule
Weather & 16 & 0.0 & 128 & 10 & 1 & $5\times e^{-5}$ & True \\
\midrule
ECL & 16 & 0.0 & 128 & 20 & 1 & $3\times e^{-4}$ & False \\
\midrule
Traffic & 16 & 0.0 & 32 & 20 & 1 & $8\times e^{-5}$ & False \\
\midrule
Solar-Energy & 8 & 1.0 & 128 & 10 & 1 & $2\times e^{-5}$ & True \\
\midrule
PEMS03 & 4 & 1.0 & 32 & 10 & 1 & $5\times e^{-5}$ & False \\
\midrule
PEMS04 & 4 & 1.0 & 32 & 5 & 1 & $5\times e^{-5}$ & False \\
\midrule
PEMS07 & 16 & 1.0 & 32 & 10 & 1 & $5\times e^{-5}$ & False \\
\midrule
PEMS08 & 16 & 1.0 & 32 & 10 & 1 & $5\times e^{-5}$ & False \\

\bottomrule

\end{tabular}

\label{tab:expsetting}
\end{threeparttable}
\end{table}
\renewcommand{\arraystretch}{1}

\renewcommand{\arraystretch}{0.95}
\begin{table*}[htb]
\setlength{\tabcolsep}{7.45pt}
\scriptsize
\centering
\begin{threeparttable}
\caption{Robustness of AMD performance. The results are obtained from five random seeds.}
\begin{tabular}{c|cc|cc|cc}
\toprule

\scalebox{1.1}{Datasets} & \multicolumn{2}{c|}{ETTh1} & \multicolumn{2}{c|}{ETTm1} & \multicolumn{2}{c}{Weather} \\ 

 \cmidrule(lr){2-3} \cmidrule(lr){4-5} \cmidrule(lr){6-7}

Metric & MSE & MAE & MSE & MAE & MSE & MAE \\
 
\midrule

96  & 0.3691$\pm$0.0008 & 0.3969$\pm$0.0001 & 0.2838$\pm$0.0004 & 0.3387$\pm$0.0003 & 0.1451$\pm$0.0003 & 0.1972$\pm$0.0002 \\
192 & 0.4008$\pm$0.0007 & 0.4160$\pm$0.0002 & 0.3218$\pm$0.0004 & 0.3618$\pm$0.0002 & 0.1868$\pm$0.0003 & 0.2381$\pm$0.0003 \\
336 & 0.4177$\pm$0.0005 & 0.4272$\pm$0.0002 & 0.3607$\pm$0.0003 & 0.3798$\pm$0.0002 & 0.2399$\pm$0.0004 & 0.2801$\pm$0.0003 \\
720 & 0.4389$\pm$0.0009 & 0.4541$\pm$0.0002 & 0.4209$\pm$0.0004 & 0.4162$\pm$0.0003 & 0.3151$\pm$0.0005 & 0.3302$\pm$0.0002 \\

\toprule

\scalebox{1.1}{Datasets} & \multicolumn{2}{c|}{Solar-Energy} & \multicolumn{2}{c|}{ECL} & \multicolumn{2}{c}{Traffic} \\
 \cmidrule(lr){2-3} \cmidrule(lr){4-5} \cmidrule(lr){6-7}
Metric & MSE & MAE & MSE & MAE & MSE & MAE \\ 
\midrule
96  & 0.1751$\pm$0.0003 & 0.2277$\pm$0.0003 & 0.1293$\pm$0.0002 & 0.2242$\pm$0.0002 & 0.3659$\pm$0.0004 & 0.2591$\pm$0.0003 \\
192 & 0.1865$\pm$0.0004 & 0.2350$\pm$0.0003 & 0.1471$\pm$0.0003 & 0.2379$\pm$0.0001 & 0.3806$\pm$0.0003 & 0.2647$\pm$0.0003 \\
336 & 0.2003$\pm$0.0003 & 0.2456$\pm$0.0002 & 0.1602$\pm$0.0003 & 0.2531$\pm$0.0002 & 0.3965$\pm$0.0004 & 0.2695$\pm$0.0003 \\
720 & 0.2032$\pm$0.0003 & 0.2490$\pm$0.0003 & 0.1928$\pm$0.0002 & 0.2860$\pm$0.0002 & 0.4292$\pm$0.0004 & 0.2918$\pm$0.0002 \\

\bottomrule

\end{tabular}
\label{tab:robust}
\end{threeparttable}
\end{table*}
\renewcommand{\arraystretch}{1}

\subsection{Hyper-Parameter Selection and Implementation Details}\label{sec:detailed parameter settings}
For the main experiments, we have the following hyper-parameters. The patch length $P$, the number of DDI blocks $n$. The dimension of hidden state of DDI $d_{model}=max\{32, 2^{[log(feature\_num)]}\}$. The number of predictor is set to 8, while the topK is set to 2. The dimension of hidden state in AMS is set to 2048. The weight decay is set to $1e^{-7}$. 
Adam optimizer is used for training and the initial learning rate is shown in \cref{tab:expsetting}.
For all datasets, to leverage more distinct temporal patterns, we set the number of MDM layers $h$ to 3 and the downsampling rate $c$ to 2. 
We report the specific hyper-parameters chosen for each dataset in \cref{tab:expsetting}. 

For all of the baseline models, we replicated the implementation using configurations outlined in the original paper or official code.

\section{Extra Experimental Results}\label{sec:extra experiment}
\subsection{Robustness Evaluation}

The results showed in \cref{tab:robust} are obtained from five random seeds on the ETTh1, ETTm1, Weather, Solar-Energy, ECL and Traffic datasets, exhibiting that the performance of AMD is stable.

\subsection{Hyper-Parameter Sensitivity}
\begin{figure*}
    \centering
    \includegraphics[width=\textwidth]{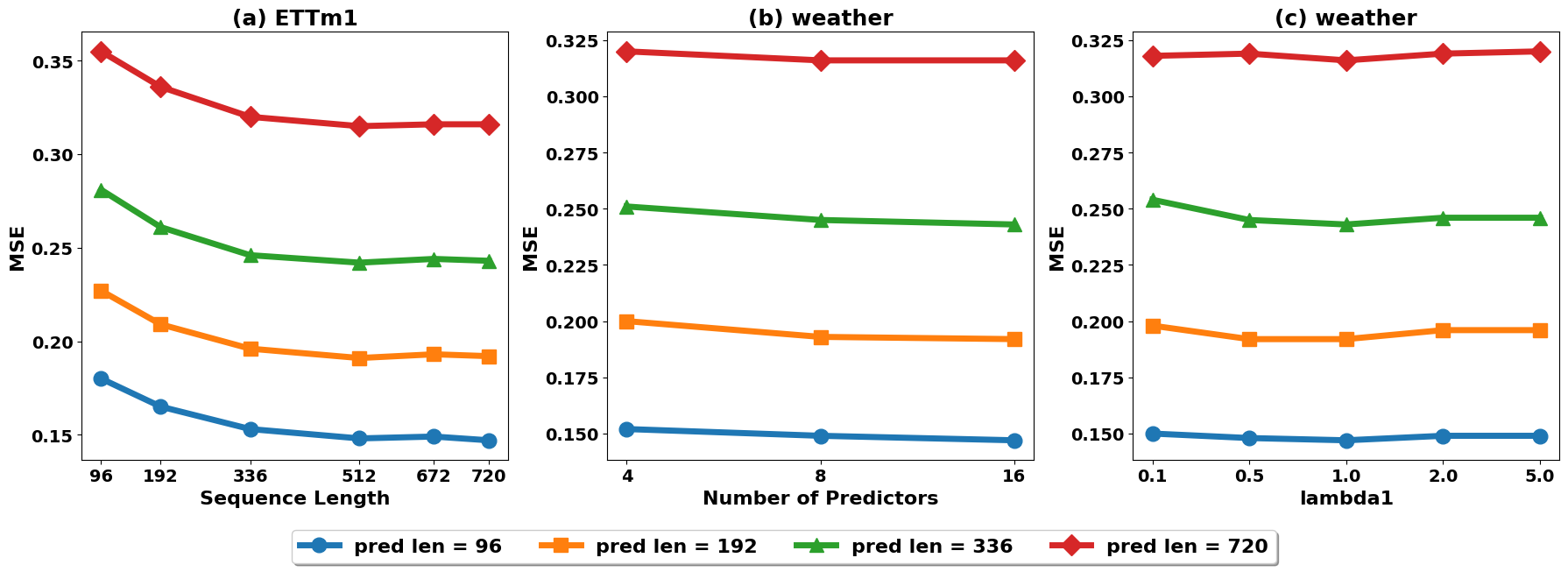} 
    \caption{Hyper-Parameter Sensitivity with respect to the input sequence length, the number of predictiors and the $\lambda_1$. The results are recorded with the all four predict length.}
    \label{fig:sensitivity}
\end{figure*}

\paragraph{Varying Input Length and Downsampling Parameters.}
Time patterns are obtained from the input sequence through downsampling. Therefore, the size of the input length $L$, the number of downsampling operations $h$, and downsampling rate $d$ all significantly affect the accuracy of prediction. To investigate the impact, we conduct the following experiments. On the ETTm1 dataset, we first choose $L$ among $\{96, 192, 336, 512, 672, 720\}$. As shown in \cref{fig:sensitivity}a, the forecasting performance benefits from the increase of input length. 
For the best input length, we choose $h$ among $\{1,2,3,4,5\}$ and $d$ among $\{1,2,3\}$. From the results shown in \cref{tab:paralayers}, it can be found that as the number of downsampling operations $h$ increases, we observe improvements in different prediction lengths. Therefore, we choose a setting of 3 layers to find a balance between efficiency and performance.

\renewcommand{\arraystretch}{0.95}
\begin{table*}[htb]
\setlength{\tabcolsep}{2.8pt}
\scriptsize
\centering
\begin{threeparttable}
\caption{The MSE resluts of different downsampling operations $h$ and downsampling rate $d$ on the ETTm1 dataset. '-' indicates no downsampling when $h=0$. In this case, $d$ has no effect on the result.}
\begin{tabular}{c|cccc|c|cccc|c|cccc}
\toprule
\begin{tabular}[c]{@{}c@{}}Predict Length\\ d=1\\ h\end{tabular} & 96 & 192 & 336 & 720 & \begin{tabular}[c]{@{}c@{}}Predict Length\\ d=2\\ h\end{tabular} & 96 & 192 & 336 & 720 & \begin{tabular}[c]{@{}c@{}}Predict Length\\ d=3\\ h\end{tabular} & 96 & 192 & 336 & 720 \\
\midrule
0 & 0.292 & 0.333 & 0.371 & 0.431 & 0 & - & - & - & - & 0 & - & - & - & - \\
1 & 0.288 & 0.326 & 0.367 & 0.428 & 1 & 0.286 & 0.326 & 0.365 & 0.427 & 1 & 0.286 & 0.327 & 0.366 & 0.428 \\
2 & 0.285 & 0.323 & 0.364 & 0.424 & 2 & 0.284 & 0.322 & 0.362 & 0.423 & 2 & 0.284 & 0.324 & 0.366 & 0.426 \\
3 & 0.285 & 0.322 & 0.362 & 0.423 & 3 & 0.284 & 0.322 & 0.360 & 0.421 & 3 & 0.283 & 0.322 & 0.363 & 0.424 \\
4 & 0.283 & 0.322 & 0.360 & 0.421 & 4 & 0.284 & 0.323 & 0.360 & 0.423 & 4 & 0.284 & 0.322 & 0.364 & 0.423 \\
5 & 0.284 & 0.323 & 0.360 & 0.422 & 5 & 0.285 & 0.321 & 0.359 & 0.423 & 5 & 0.284 & 0.322 & 0.363 & 0.422 \\

\bottomrule

\end{tabular}
\label{tab:paralayers}
\end{threeparttable}
\end{table*}
\renewcommand{\arraystretch}{1}

\paragraph{Varying Number of Predictors and TopK.}
The number of decomposed temporal patterns and the dominant temporal patterns are determined by the number of predictors and the topK value, respectively. We conduct hyperparameter sensitivity experiments about these two parameters on the Weather dataset and the results are shown in \cref{fig:sensitivity}b. We observe an improvement in prediction results as the number of predictors increases. However, this also leads to an increase in the memory usage of selector weights. To strike a balance between memory and performance, we finally opt for 8 predictors.

\paragraph{Varying Scaling of $\mathcal{L}_\text{selector}$.}
We conduct an investigation into the scaling factor for the load balancing loss $\mathcal{L}_\text{selector}$, denoted by $\lambda_1$, on the Weather dataset. As shown in observation 4, we found that when $\lambda_1$ is set to $0$, the loss function does not guide the reasonable assignment of different temporal patterns to different predictors, resulting in poor prediction performance. However, when $\lambda_1$ is not $0$, as shown in the \cref{fig:sensitivity}c, the model demonstrates strong robustness to $\lambda_1$, with prediction results independent of scaling. Upon observing the loss function, we noticed that $\mathcal{L}_\text{selector}$ has already decreased significantly in the early epochs. Therefore, we finally choose $\lambda_1$ to be $1.0$.

\begin{figure*}
    \centering
    \includegraphics[width=\textwidth]{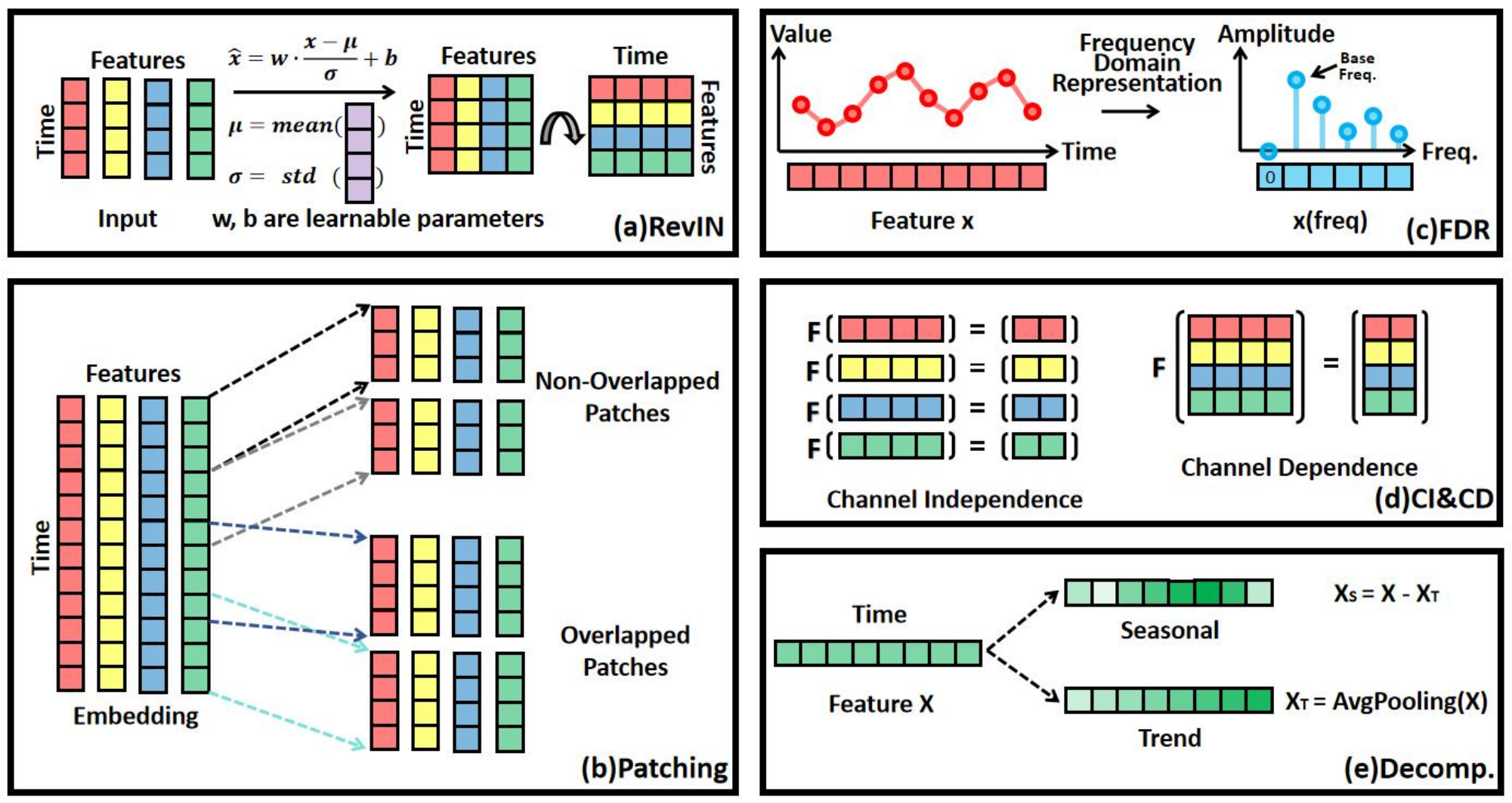} 
    \caption{Five Designs for Sequential Modelling  with respect to the RevIN, Patching, Frequency Domain Representation, Channel Independence and Sequence Decomposition.
    }
    \label{fig:design}
\end{figure*}

\section{Discussions on Limitations and Future Improvement}\label{sec:limitations ans future improvement}

Recently, several specific designs have been utilized to better capture complex sequential dynamics, such as normalization, patching, frequency domain representation, channel independence, sequence decomposition, and others, as shown in \cref{fig:design}.

(1) Normalization : Real-world time series always exhibit non-stationary behavior, where the data distribution changes over time. 
RevIn~\cite{revin} is a normalization-and-denormalization method for TSF to effectively constrain non-stationary information (mean and variance) in the input layer and restore it in the output layer. The work has managed to improve the delineation of temporal dependencies while minimizing the influence of noise. In AMD, we also adopt the method of Revin. 
However, it struggles to adequately resolve the intricate distributional variations among the layers within deep networks, so we need to make further improvements to address this distributional shift.

(2) Patching : Inspired by the utilization of local semantic context in computer vision (CV) and natural language processing (NLP), the technique of patching is introduced~\cite{patchtst}. Since TS data exhibit locality, individual time steps lack the semantic meaning found in words within sentences. Therefore, extracting the local semantic context is crucial for understanding their connections. Additionally, this approach has the advantage of reducing the number of parameters. In AMD, we also develop a patching mechanism to extract recent history information. However, how to better exploit locality remains an issue that requires further research. 

(3) Frequency Domain Representation : TS data, characterized by their inherent complexity and dynamic nature, often contain information that is sparse and dispersed across the time domain. The frequency domain representation is proposed to promise a more compact and efficient representation of the inherent patterns. Related methods~\cite{fedformer} still rely on feature engineering to detect the dominant period set. However, some overlooked periods or trend changes may represent significant events, resulting in information loss. In the future, we can explore adaptive temporal patterns mining in the frequency domain, thereby utilizing the Complex Frequency Linear module proposed by FITS~\cite{fits} to mitigate the problem of large parameter sizes in MLP-based models when the look-back length is long.

(4) Channel Independence (CI) and Channel Dependence (CD) : CI and CD represent a trade-off between capacity and robustness~\cite{civscd}, with the CD method offering greater capacity but often lacking robustness when it comes to accurately predicting distributionally drifted TS. In contrast, the CI method sacrifices capacity in favor of more reliable predictions. PatchTST~\cite{patchtst} achieves SOTA results using the CI approach. However, neglecting correlations between channels may lead to incomplete modelling. In AMD, we leverage the CI approach while integrating dependencies across different variables over time, thus exploiting cross-channel relationships and enhancing robustness. However, the trade-off between capacity and robustness is also a balance between generalization and specificity, which still requires further research.

(5) Sequence Decomposition : The classical TS decomposition method~\cite{HOLT} divides the complex temporal patterns into seasonal and trend components, thereby benefiting the forecasting process. In AMD, we go beyond the constraints of seasonal and trend-based time series decomposition. Instead, we develop an adaptive decomposition, mixing and forecasting module that fully exploits the information from different temporal patterns. However, the effectiveness of the adaptive decomposition module relies heavily on the availability and quality of historical data, which poses challenges in scenarios with limited or noisy data.

We believe that more effective sequence modelling designs will be proposed to adequately address issues such as distribution shift, multivariate sequence modelling, and so on. As a result, MLP-based models will also perform better in more areas of time series.

\end{document}